\documentclass[preprint,12pt]{elsarticle}

\journal{Computer Physics Communications}
\usepackage{amssymb}
\usepackage{amsmath}
\usepackage{physics}
\usepackage{cleveref}
\usepackage{mathrsfs}
\usepackage{xspace}
\usepackage{subfig}
\usepackage{tikz}
\usepackage{todonotes}
\usepackage{lineno} 
\usepackage{xcolor}

\def\*#1{\mathbf{#1}}
\def\^#1{\widehat{#1}}
\def\-#1{\overline{#1}}
\def\~#1{\widetilde{#1}}
\def\>#1{\overrightarrow{#1}}

\DeclareMathOperator*{\argmax}{arg\,max}
\DeclareMathOperator*{\argmin}{arg\,min}

\newcommand{\rpde}{\mathcal{R}_{\text{PDE}}}
\newcommand{\nn}{\mathcal{M}_{\theta}}
\newcommand{\ourmethod}{\texttt{DeepBayONet}\xspace}

\newenvironment{highlightblue}
  {\begingroup                 
  }
  {\endgroup}      
  
\newif\iftechreport
\techreporttrue
\begin{document}

\begin{frontmatter}



\title{Deep Operator Networks \\ for Bayesian Parameter Estimation in PDEs}

\author[csulb]{Amogh Raj}
\ead{amogh@csml-research.org}

\author[csulb]{Sakol Bun} 
\ead{sbun@csml-research.org}

\author[csulb]{Keerthana Srinivasa} 
\ead{keerthana@csml-research.org}

\author[csulb]{Carol Eunice Gudumotou} 
\ead{caroleunicetr@gmail.com}

\author[csulb]{Arash Sarshar\corref{cor1}} 
\cortext[cor1]{Corresponding author \texttt{(arash.sarshar@csulb.edu)}}

\affiliation[csulb]{organization={Department of Computer Engineering and Computer Science},
addressline={ California State University},
city={Long Beach},
postcode={90840},
state={CA},
country={USA}}

\begin{abstract}
We present a novel framework combining  Deep Operator Networks (DeepONets) with Physics-Informed Neural Networks (PINNs) to solve partial differential equations (PDEs) and estimate their unknown parameters. By integrating data-driven learning with physical constraints, our method achieves robust and accurate solutions across diverse scenarios. Bayesian training is implemented through variational inference, allowing for comprehensive uncertainty quantification for both data and model uncertainties. This ensures reliable predictions and parameter estimates even in noisy conditions or when some of the physical equations governing the problem are missing. The framework demonstrates its efficacy in solving forward and inverse problems, including the 1D unsteady heat equation, 2D reaction-diffusion equations,  3D eigenvalue problem, and various regression tasks with sparse, noisy observations. This approach provides a computationally efficient and generalizable method for addressing uncertainty quantification in PDE surrogate modeling. 
\end{abstract}

\iftechreport
\fi


\end{frontmatter}



\section{Introduction}
\label{sec:introduction}

The modeling and simulation of complex physical systems governed by partial differential equations (PDEs) are critical in science and engineering. Traditional numerical methods for solving PDEs, while effective, often demand substantial computational resources and face challenges with high-dimensional or nonlinear problems \cite{surrogateML,sarshar2017numerical}. Neural networks have emerged as promising tools for approximating PDE solutions \cite{chen2019optimal,hwang2021solving,sifan2021deeponet,bhattacharjee2024improving}. However, these models typically require large datasets and lack mechanisms to embed physical laws into their frameworks.
    
Physics-Informed Neural Networks (PINNs) \cite{pinns2019Karniadakis,lu2021deepxde} integrate governing physical laws, expressed as PDEs, directly into their loss functions. This approach enables PINNs to solve forward and inverse problems efficiently, even with limited data, by ensuring that predictions adhere to the underlying physics. By embedding physical laws into the learning process, PINNs overcome the reliance on extensive datasets required by traditional neural networks, making them more practical for real-world scenarios \cite{pinn2022Cuomo}. Applications of PINNs include solving fluid dynamics problems \cite{pinnFlow2020Mao,pang2019},  inverse problems \cite{chen2020,QiZhi2020}, controlling dynamical systems \cite{antonelo2021physicsinformed,barry2022physics,nellikkath2021physicsinformed}, and performing uncertainty quantification \cite{zhang2019}.

Automatic differentiation (AD) is integral to the functionality of PINNs, enabling precise and efficient computation of PDE residuals, even for mixed and high-order derivatives. Its integration into machine learning frameworks such as \texttt{PyTorch} \cite{adPytorch2017paszke}, \texttt{Jax} \cite{jax2018github}, and \texttt{TensorFlow} \cite{tensorflow2015-whitepaper} has significantly advanced deep learning applications in scientific domains. Comprehensive surveys on AD and its relationship to deep learning can be found in Baydin et al. \cite{ad2018Baydin} and Margossian \cite{ad2019Margossian}.

In scientific machine learning, probabilistic models are highly valued for estimating noise in observational data, handling model errors, and accounting for incomplete knowledge of the underlying physics. Viana et al. \cite{bpinn2021Viana} highlight the evolution of scientific modeling, emphasizing the role of physics-informed neural networks in reducing computational costs and enhancing modeling flexibility. Molnar et al. \cite{bpinn2022Molnar} demonstrate the application of PINNs within a Bayesian framework to reconstruct flow fields from projection data, showcasing their ability to incorporate physics priors and improve accuracy in noisy environments. Li et al. \cite{bpinn2023Li} explore the challenges of solving inverse problems with PINNs, advocating a Bayesian approach to provide robust uncertainty quantification, particularly when observations are sparse or noisy.

Bayesian Neural Networks (BNNs) \cite{bnn2020wang,bnn2022Bonneville,bnn2021Izmailov} provide a probabilistic framework for uncertainty quantification by assigning distributions to network weights. While posterior estimation over BNN parameters is challenging due to high-dimensionality and non-convexity, methods such as Hamiltonian Monte Carlo (HMC) and variational inference (VI) approximate the posterior effectively, albeit at higher computational costs than maximum likelihood approaches. Yang et al. \cite{bpinns2021yang} introduced Bayesian Physics-Informed Neural Networks (B-PINNs), combining BNNs and PINNs to address forward and inverse PDE problems in noisy conditions. By integrating physical equations into training, B-PINNs offer a practical solution to probabilistic PDE learning tasks.

\begin{highlightblue}
Accurately estimating PDE parameters while quantifying uncertainty remains a significant challenge, particularly in complex systems with scarce or noisy data. While Deep Operator Networks \cite{deeponet2019lu, sifan2021deeponet, lu2021deepxde} can learn PDE solution operators from data, they provide deterministic predictions and lack a principled mechanism for uncertainty quantification. To resolve this limitation, B-DeepONet \cite{LIN2023BDEEPONET} extends the DeepONet framework by introducing a Bayesian formulation, enabling posterior sampling over the solution operator space. B-DeepONet learns model uncertainty by learning a posterior distribution over model parameters using Langevin dynamics. During training, a variant of Stochastic Gradient Langevin Dynamics (SGLD) applies gradients alternatively to the parameters of the branch and trunk networks to help reduce the computational cost of calculating gradients for the ensemble of parameters.
\end{highlightblue}

The high computational demands and challenges in selecting suitable priors in Bayesian approaches restrict the practical use of black-box sampling methods in scientific machine learning. This presents an opportunity for new methodologies that effectively combine physics-guided training with computationally efficient Bayesian approaches.

In this work, we propose a novel method that extends the capabilities of PINNs by integrating Bayesian methodologies with Deep Operator Networks (DeepONets) \cite{lu2019deeponet} to improve parameter estimation in PDEs. Our approach relies on variational inference: starting with a prior distribution for the unknown PDE parameters, we apply inference to estimate the posterior distribution of the PDE solution and the parameters based on the available data and known physics of the problem. The DeepOperator architecture modulates the forward prediction of the solution of the PDE with inverse parameter estimation. This method offers flexible uncertainty quantification and accurate differentiable solutions.

We continue our discussion in the following steps: In \cref{sec:methodology}, we detail the methodology of Bayesian Deep Operator Networks, demonstrate their application to parameter estimation in PDEs, and discuss practical ways of quantifying model and data uncertainty in their predictions. \Cref{sec:experiments} presents experimental results on a variety of PDEs, which showcase the effectiveness of our method in solving forward and inverse problems. Finally, we discuss the implications of our findings for future research and potential applications in various scientific and engineering domains in \cref{sec:conclusions}. 
\begin{highlightblue}
    The source code for these experiments is publicly available as Jupyter Notebooks at \texttt{https://github.com/csml-beach/deep-bayesian-operator-nets}.
\end{highlightblue}
\section{Methodology}
\label{sec:methodology}

\subsection{Problem definition}

We start by considering a partial differential equation (PDE) with a latent parameter $\lambda$ defined over the time span $t \in [t_0, t_f ]$ 
and spatial domain $x \in \Omega$:
\begin{equation}
        {F}(t,x,y; \lambda ) = 0, \quad (t,x) \in [t_0, t_f ] \times \Omega.
        \label{eq:Forward-PDE}
\end{equation}
${F}$ may contain partial derivatives of the solution of the forms $\pdv[\alpha]{y}{x}$ and $\pdv[\beta]{y}{t}$. $y(t,x; \lambda)$ 
is the forward solution of the PDE and is assumed to be well defined with the necessary initial and boundary conditions. 
Given a dataset of noisy observations of $y$ at different times and positions: 
\begin{equation}
    \mathcal{D} = \{(t_i, x_i, y_i), ~ i=0, 1,\ldots, N\}, \label{eq:data}
\end{equation}
we are interested in training a probabilistic machine learning model that simultaneously approximates $y(t,x)$ and $\lambda$ based on the data \cref{eq:data} and the PDE \cref{eq:Forward-PDE}.  We will start from a Bayesian formulation of this problem and derive a corresponding variational loss function for training the model.
%
%

The posterior distribution over the latent parameter  given the data  is expressed as
\begin{equation}
p(\lambda \mid \mathcal{D}) = \frac{p(\mathcal{D} \mid \lambda)\, p_0(\lambda)} {p(\mathcal{D})}, \label{eq:Bayes}
\end{equation}
where $p_0(\lambda)$ represents the prior distribution over $\lambda$, $p(\mathcal{D} \mid \lambda)$ is the likelihood of the observed data given parameter $\lambda$, and $p(\mathcal{D})$ is the marginal likelihood or evidence.

The likelihood term $p(\mathcal{D} \mid \lambda)$ is constructed from two components: i) the data likelihood $p(y_i \mid t_i, x_i, \lambda)$ captures the relationship between the inputs $t_i, x_i$ and outputs $y_i$. ii) and the likelihood of the predicted solution satisfying the governing equations. The latter is calculated based on the residuals of \cref{eq:Forward-PDE} when the model predicts the parameter and the solution. We refer to this residual as $\rpde$ hereafter. 


Assuming independence across training data points, the joint likelihood is written as
\begin{equation}
P(\mathcal{D} \mid \lambda) = \prod_{i=1}^N p(y_i \mid t_i, x_i, \lambda) \, p(\rpde(\nn(x_i, \lambda))).
\end{equation}

The data likelihood $p(y_i \mid x_i, \lambda)$ is modeled with normal distribution:
\begin{equation}
p(y_i \mid t_i, x_i, \lambda) = \mathcal{N}(y_i; \nn(t_i, x_i, \lambda), \sigma_y^2),
\end{equation}
where $\nn(t_i, x_i, \lambda)$ is the model output for input $(t_i, x_i)$ given latent variable $\lambda$, and $\sigma_y^2$ is the observation noise variance. The log-likelihood is therefore
\begin{equation}
\log p(y_i \mid t_i, x_i, \lambda) = -\frac{|\nn(t_i, x_i, \lambda) - y_i|^2}{2\sigma_y^2} - \frac{1}{2} \log(2\pi\sigma_y^2). \label{eq:log-likelihood-data}
\end{equation}

We are interested in models that, in addition to adherence to observation data, also produce likely predictions that satisfy the governing equations \cref{eq:Forward-PDE}.  The likelihood of the residuals from the PDE is similarly modeled as:
\begin{equation}
p(\rpde(\nn(t_i, x_i, \lambda))) = \mathcal{N}(0, \sigma_R^2), \label{eq:residual-likelihood}
\end{equation}
where $\sigma_R^2$ is the variance of the residual noise. $\rpde$ can either be simplified as $ \sum_i |F(t_i, x_i, y_i, \lambda)|$ or modified to include other constraints such as the boundary and initial conditions. The corresponding log-likelihood of the residuals is
\begin{equation}
\log p(R_{\text{PDE}}(\nn(x_i, \lambda))) = -\frac{|\rpde(\nn(t_i, x_i, \lambda))|^2}{2\sigma_R^2} - \frac{1}{2} \log(2\pi\sigma_R^2). \label{eq:log-likelihood-residual}
\end{equation}

We sample the posterior $p(\lambda \mid \mathcal{D})$ in equation \cref{eq:Bayes} using variational inference (VI). Considering an approximate posterior distribution $q(\lambda)$, we want to minimize the Kullback-Leibler (KL) divergence:
\begin{equation}
\argmin_{q(\lambda)} D_{KL}(q(\lambda) \| p(\lambda \mid \mathcal{D})) = \int q(\lambda) \log \frac{q(\lambda)}{p(\lambda \mid \mathcal{D})} d\lambda.
\end{equation}
Substituting \cref{eq:Bayes} for $p(\lambda \mid \mathcal{D})$ and rearranging terms, yields the well-known Evidence Lower Bound (ELBO) objective function:
\begin{equation}
\argmax_{q(\lambda)}\mathcal{L}_{\text{ELBO}} = \int q(\lambda) \log \frac{p(\mathcal{D} \mid \lambda) p_0(\lambda)}{q(\lambda)} \dd{\lambda}. \label{eq:ELBO-primiary}
\end{equation}

\begin{highlightblue}
Defining the expectation on $ \lambda \sim q(\lambda)$ :
\[
\mathbb{E}_{q(\lambda)}[\,f(\lambda)\,] := \int f(\lambda) \; q(\lambda) \, d\lambda,
\]
\end{highlightblue}
 and separating $p(\mathcal{D} \mid \lambda)$ using the joint likelihood, \cref{eq:ELBO-primiary} becomes
    \begin{multline}
\mathcal{L}_{\text{ELBO}} = \mathbb{E}_{q(\lambda)} \left[ \sum_{i=1}^N \log p(y_i \mid x_i, \lambda) + \log p(\rpde(\nn(x_i, \lambda))) \right] \\ - D_{KL}(q(\lambda) \| p_0(\lambda)). \label{eq:ELBO-secondary}
    \end{multline}


Finally, substituting the log-likelihoods \cref{eq:log-likelihood-data,eq:log-likelihood-residual}, the loss function becomes
\begin{align}
\mathcal{L} = \mathbb{E}_{q(\lambda)} \Bigg[ & \sum_{i=1}^N \left( -\frac{|\nn(t_i, x_i, \lambda) - y_i|^2}{2\sigma_y^2} - \frac{1}{2} \log(2\pi\sigma_y^2) \right)\notag \\
& - \frac{|\rpde(\nn(t_i, x_i, \lambda))|^2}{2\sigma_R^2} - \frac{1}{2} \log(2\pi\sigma_R^2) \Bigg] - D_{KL}(q(\lambda) \| p_0(\lambda)). \label{eq:ELBO-final}
\end{align}

\Cref{eq:ELBO-final} can be evaluated using Monte Carlo sampling of the latent variable $\lambda$ from the posterior $q(\lambda)$. 
The observation and residual noise variances $\sigma_y^2$ and $\sigma_R^2$ can be considered constants or learned during training. This enables a flexible uncertainty quantification method that can be utilized to estimate model or data uncertainty. We provide examples for different uncertainty estimation strategies in the experiments \cref{sec:experiments}.

The KL divergence term in \cref{eq:ELBO-final}  can be approximated empirically or analytically: 
If both the prior $p_0(\lambda)$ and the variational distribution $q(\lambda)$ are assumed to be normally distributed, with means $(\mu_0 = 0, \mu_q)$ and variances $(\sigma_0^2 =1 , \sigma_q^2)$, the KL divergence can be written as:
\begin{equation}
D_{KL}(q(\lambda) \| p_0(\lambda)) = - \log {\sigma_q} + \frac{\sigma_q^2 + \mu_q^2 }{2} - \frac{1}{2}.
\end{equation}

\subsection{Model architecture}
\begin{figure}
    \centering
    \includegraphics[width=0.99\textwidth]{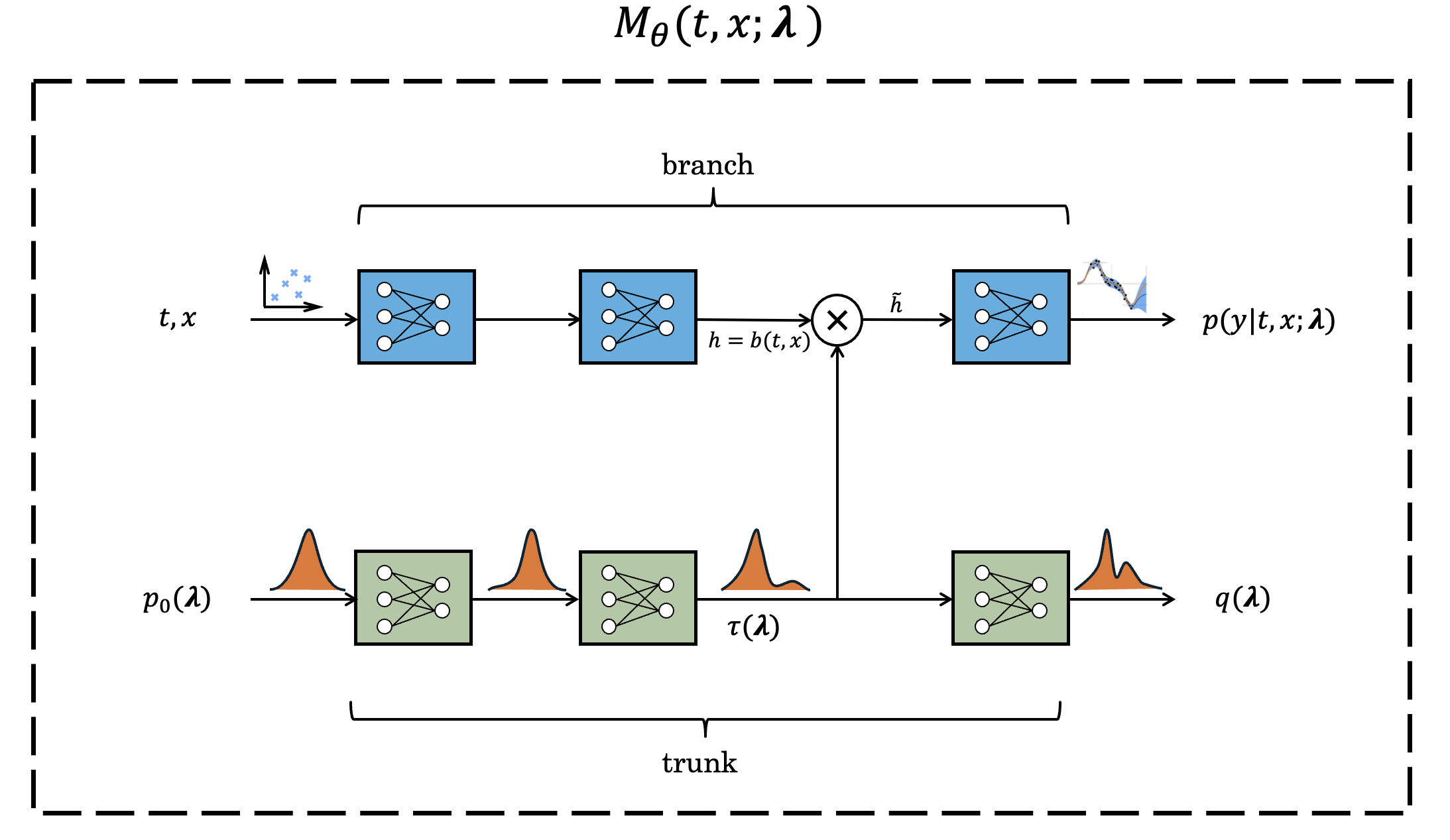}
    \captionsetup{justification=centering}
    \caption{\ourmethod feedforward architecture with trunk and branch networks. The trunk network learns the PDE parameters, while the branch network predicts the forward solution of the problem.}
    \label{fig:architecture}
\end{figure}

We used a deep operator network (DeepONet) \cite{chen95,lu2019deeponet} to parametrize $\nn(t,x;\lambda)$. This architecture consists of two components: the branch network and the trunk network.  The branch network approximates the forward solution $y(t,x)$ of the PDE. The trunk network processes independent prior samples from a standard normal distribution and transports them to samples of the approximate posterior $q(\lambda)$. The trunk and branch exchange information using an element-wise product before the output layers \cite{deeponet2019lu}. See \cref{fig:architecture} for a schematic representation of different components of the architecture. Our proposed network, referred to as \ourmethod, uses the DeepONet architecture to estimate the forward solution and parameters of the PDE at the same time:
\begin{equation}
    y(t,x; \lambda) \approx \nn(t,x,\lambda) = \sigma \Big( W_0 \big( \mathsf{b}(t, x) \odot \mathsf{\tau}(\lambda) \big) + b_0 \Big),
    \label{eq:DeepONet}
\end{equation}
where $W_0$ and $b_0$ are the output layer weights and bias, $\sigma(\cdot)$ is the activation function, and $\odot$ denotes the element-wise product. The branch network $\mathsf{b}(t,x)$ and trunk network $\mathsf{\tau}(\lambda)$ are general parametrized functions and can be implemented using any neural network architectures, including convolutional, recurrent, or transformer networks. The entire set of trainable parameters in the model is denoted by $\theta$.
\subsection{ DeepBayONet Uncertainty estimation}

Uncertainty estimation is critical in deep neural networks, particularly in scientific modeling and prediction tasks, to assess the reliability of predictions and the robustness of the model. Two sources of uncertainty can be considered: data uncertainty arises from inherent noise in the observations, reflecting variability that cannot be reduced even with additional data. Under normal distribution assumptions, this type of uncertainty is often captured by modeling the observation error (co)variance. Specifically in deep learning, the predicted output of the network is augmented to include both the mean-prediction as well as the variance of the predicted output \cite{kendall_what_2017}.

In contrast, model uncertainty reflects the network's lack of flexibility or insufficient training around the underlying data distribution and can be reduced by incorporating additional data or improving the model architecture. Bayesian neural networks (BNNs) and dropout are among various methods that can address epistemic uncertainty \cite{kendall_what_2017} by learning a distribution over the model parameters, allowing the model to express uncertainty in regions of the input space where the model's predictions are ambiguous. These two sources of uncertainty provide complementary information: data uncertainty quantifies noise in the data, while model uncertainty assesses the model's confidence in its predictions.

In \ourmethod we have the ability to quantify both sources of uncertainty. Data uncertainty can be captured through a learnable observation noise variance \(\sigma_y^2(x;\theta)\) which appears in the data likelihood term of the ELBO loss \cref{eq:ELBO-final} and accounts for the inherent variability in the measurements. By treating \(\sigma_y^2\) as a learnable parameter, the model can adaptively estimate the data uncertainty based on the inputs. High values of \(\sigma_y^2\) indicate regions with significant observation noise, leading to reduced confidence in the predictions.

\begin{highlightblue}
Model uncertainty in \ourmethod is inferred by randomly perturbing the latent representation $h(t,x;\lambda) = b(t,x; \lambda) $ of the forward solution. This latent representation is taken from the deep layers in the branch network before the output layer (see \cref{fig:architecture}).  This mechanism of perturbation bears a conceptual resemblance to dropout \cite{srivastava2014dropout}, widely used for regularization as well as epistemic (model) uncertainty quantification. Dropout introduces noise during training by randomly setting activations to zero with a probability sampled from a Bernoulli distribution \cite{srivastava2014dropout}. Let ${h} \in \mathbb{R}^d$ denote the hidden activations of a layer. During dropout, each element $h_i$ is multiplied by a binary random variable $z_i$ drawn independently from a Bernoulli distribution:

\begin{subequations}
\begin{align}
z_i &\sim \mathrm{Bernoulli}(p), \quad \text{for } i = 1, \dots, d, \\
\tilde{{h}} &= {z} \odot {h},
\label{eq:dropout}
\end{align}
\end{subequations}
where $p \in (0, 1)$ is the retention probability and $ z = [z_1, \dots, z_d]^\top$ is the dropout mask. This formulation reveals that dropout is equivalent to injecting multiplicative noise sampled from a discrete $\{0, 1\}$ distribution into the network during training.

Our framework adopts a similar stochastic perturbation principle to that of dropout, but with a key difference:  \ourmethod introduces \emph{continuous perturbations} to the solution representation. These perturbations are drawn from a distribution over perturbations learned by the trunk network:

\begin{subequations}
\begin{align}
\lambda &\sim \mathcal{N}(0,\mathbf{I}), \\
\tilde{{h}} &= \tau(\lambda) \odot {h}.
\label{eq:deepoUQ}
\end{align}   
\end{subequations}

This continuous perturbation mechanism has two significant implications. First, it ensures that the model's predictions generalize well by preventing the branch network from overfitting to deterministic solutions. Second, it provides a means to quantify the discovered parameters' uncertainty, as the variability in \(\mathsf{\tau}(\lambda)\) represents the model's confidence in its parameter estimates. The posterior distribution \(q(\lambda)\) over the latent parameters can be empirically estimated using the trunk network to understand the model's confidence in the predicted parameter.

Finally, we note that if one is interested in estimating the residual noise variance \(\sigma_R^2\), associated with the residuals of the governing equation \cref{eq:log-likelihood-residual}
It can also be learned as a parametric function of the network.

The \ourmethod{} architecture naturally integrates regularization and uncertainty quantification in a unified framework, leveraging the interaction between the branch and trunk networks to achieve both objectives.

\end{highlightblue}




\subsection{Practical Aspects of the Loss Function}

\begin{highlightblue}
    
It is possible to numerically optimize the cost function \cref{eq:ELBO-final} through Monte Carlo sampling over $t,x,\lambda$.  When input samples are drawn from the training data, the data likelihood term will guide the model to predict the known forward solution. When inputs are not in the dataset, the residual likelihood trains the model to adhere to physical constraints and the PDE equations. Simultaneously, the entropy term in \cref{eq:ELBO-final} encourages the posterior of the learned parameter to remain normally distributed.  

A common practice in the PINNs literature is to scale different components of the loss function to give importance to certain aspects and de-emphasize others.  We follow the same practical technique and assign different weights to the terms in \cref{eq:ELBO-final}.

 Assuming \(\nn(t_i, x_i; \lambda) = \tilde{y}_i\) is the predicted PDE solution, the total loss is expressed as:
\begin{highlightblue}
    
\begin{equation}
\mathcal{L} = w_{\text{Interior}} L_{\text{Interior}} + w_{\text{IC}} L_{\text{IC}} + w_{\text{BC}} L_{\text{BC}} + w_{\text{STD}} L_{\text{STD}} + w_{\text{Data}} L_{\text{Data}},
 \label{eq:total_loss_practical}
\end{equation}
\end{highlightblue}

where
\begin{subequations}
\begin{align}
    L_{\text{Interior}}(\theta, \lambda; X_{\text{Interior}}) &= \frac{1}{|X_{\text{Interior}}|} \sum_{x \in \Omega} \frac{\left\| F(t, x, y; \lambda) \right\|^2}{\sigma_R^2}, \label{eq:interior_loss_practical} \\
    L_{\text{BC}}(\theta, \lambda; X_{\text{BC}}) &= \frac{1}{|X_{\text{BC}}|} \sum_{x \in \partial \Omega} \frac{\left\| \nn(t, x; \lambda) - y(t, x; \lambda) \right\|^2}{\sigma_y^2}, \label{eq:bc_loss_practical} \\
    L_{\text{IC}}(\theta, \lambda; X_{\text{IC}}) &= \frac{1}{|X_{\text{IC}}|} \sum_{x \in \Omega} \frac{\left\| \nn(0, x; \lambda) - y(0, x; \lambda) \right\|^2}{\sigma_y^2}, \label{eq:ic_loss_practical} \\
    L_{\text{Data}}(\theta, \lambda; X_{\text{Data}}) &= \frac{1}{|X_{\text{Data}}|} \sum_{(x) \in X_{\text{Data}}} \frac{\left\| \nn(t, x; \lambda) - y(t, x) \right\|^2}{\sigma_y^2}, \label{eq:data_loss_practical} \\
    L_{\text{STD}}(\theta, \lambda) &= \left\| \log{\sigma_y^2} \right\|,\label{eq:std_loss_practical}
\end{align}
\end{subequations}
where we have expanded the log-likelihood of the data term in \cref{eq:ELBO-final} to also include any boundary and initial conditions we may want to enforce for the learned solution.\footnote{\footnotesize {The loss component in \cref{eq:std_loss_practical} was found empirically to be simpler than evaluating the exact KL-divergence. The intuition behind it is that the log of the variance of the predicted solution needs to be of moderate magnitude. Too large and too small variances are discouraged.}}

The weights \(w_{\text{Interior}}, w_{\text{IC}}\), etc.,  are inevitably hyperparameters that balance the influence of the respective loss components, and their choice can highly affect the quality of the learned solution. Adaptively choosing these weights is an active area of research in PINNs \cite{xiang2022self,mcclenny2023self}. In this paper, we have used a combination of grid search and hand-tuning of hyperparameters.

\end{highlightblue}
\section{Experiments}
\label{sec:experiments}
In this section, we train \ourmethod{s} on various algebraic and differential equations. In many cases, the underlying equations have unknown parameters that are learned along with the forward dynamics. We compare the results with exact solutions and competing methodologies from the literature when they are available.

\subsection{General experiment setup}
\label{subsec:general_setup}
\begin{highlightblue}
The common configurations used for all the experiments are as follows: each experiment utilizes a feedforward neural network architecture comprising three layers on both the branch and the trunk. The hyperbolic tangent (\(\tanh\)) activation function is employed universally. Training and optimization are conducted using the \texttt{Adam} optimizer. All implementations use the \textit{PyTorch} package, leveraging its features for computational graphs and gradients. Specific details, such as the number of neurons per layer, epochs, batch size, and loss weights, are tuned for each experiment based on its requirements and complexity. These experiments are documented in the paper's GitHub repository\footnote{\texttt{https://github.com/csml-beach/deep-bayesian-operator-nets}} for reference. 
\end{highlightblue}

\subsection{Forward Bayesian Regression Problem}
\label{subsec:forward_bayesian}
\begin{figure}[h]
    \centering
    \subfloat[]{
        \includegraphics[width=0.31\textwidth]{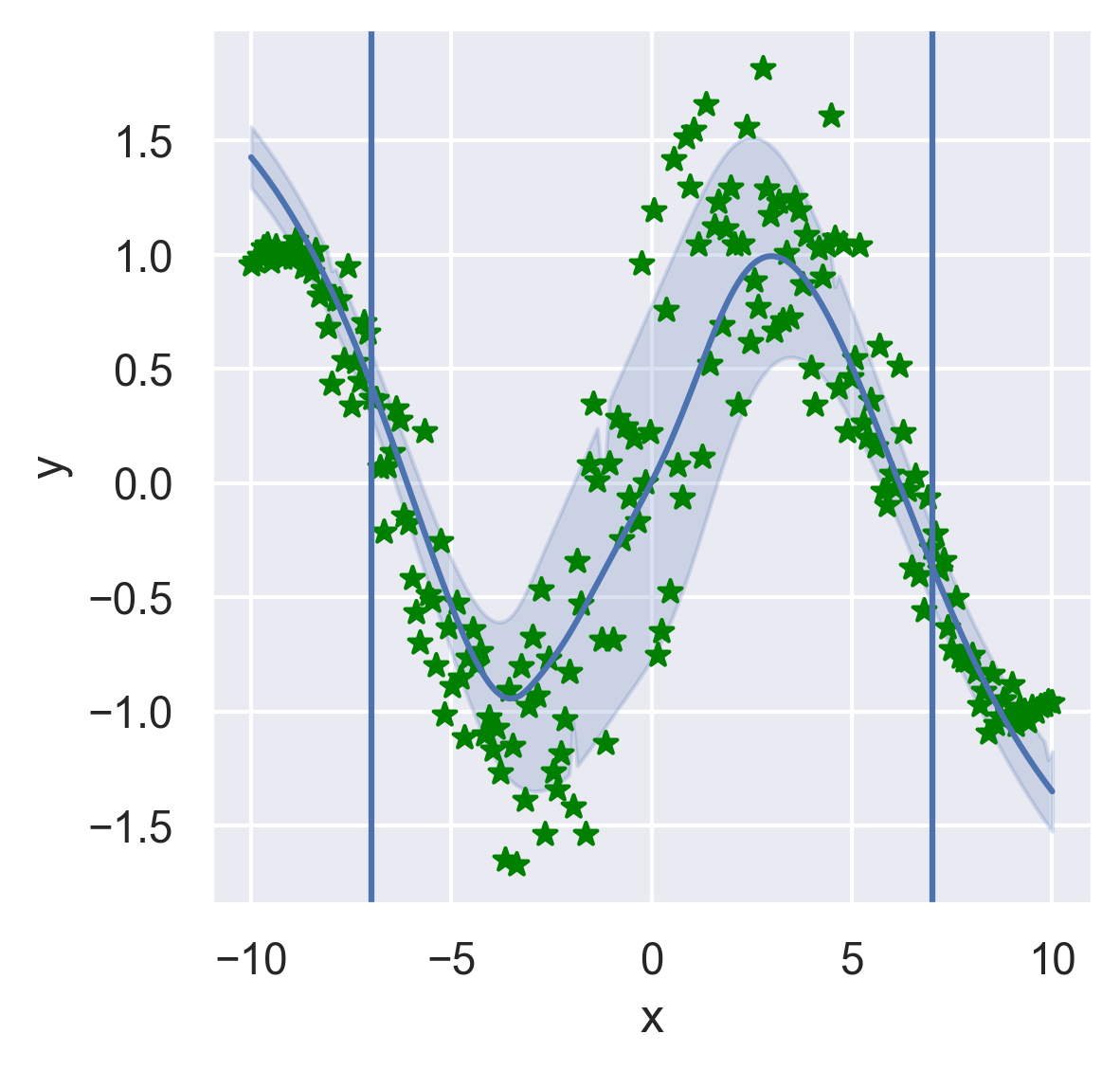}
        \label{fig:snn}
    }
    \subfloat[]{
        \includegraphics[width=0.31\textwidth]{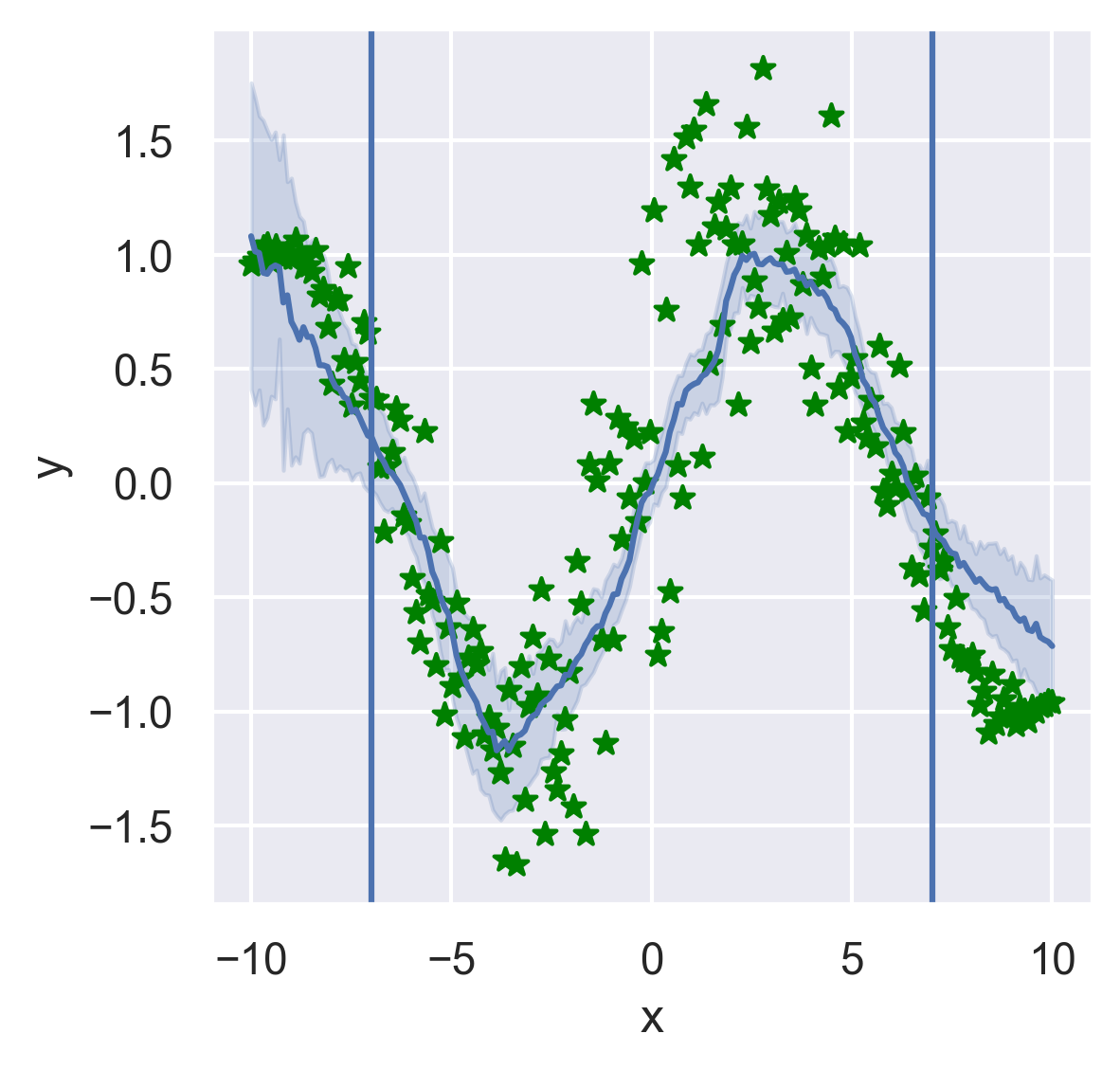}
        \label{fig:mcd}
    }
    \subfloat[]{
        \includegraphics[width=0.31\textwidth]{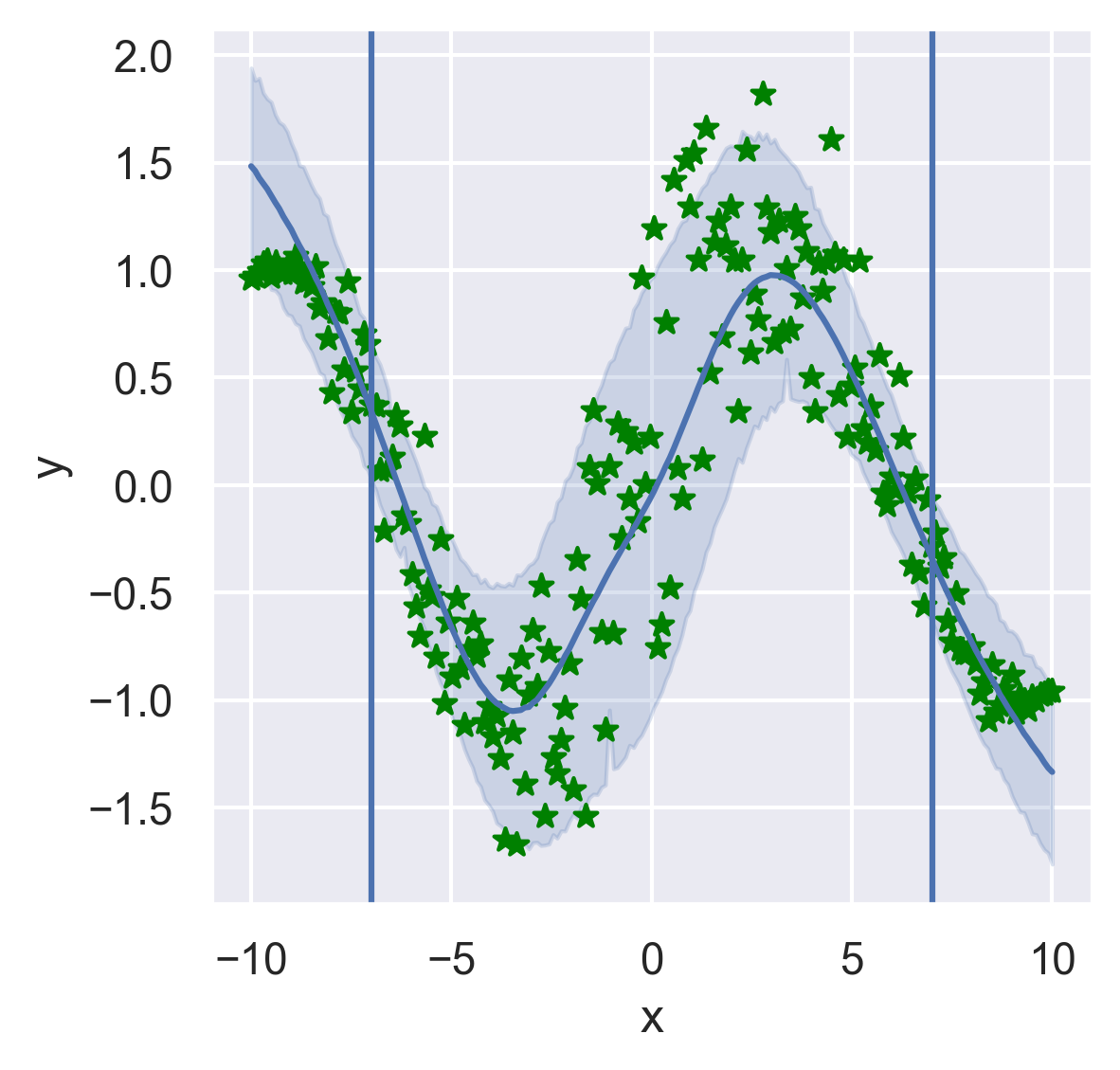}
        \label{fig:dpo}
    }
    \caption{(a) DeepBayONet without trunk (SNN). (b) Monte Carlo Dropout (MCDO). (c) Deep Bayesian Operator Network (DeepBayONet). Each method exhibits distinct uncertainty quantification behavior under the same experimental conditions. The two vertical blue lines in each plot denote the division between the IDD dataset (between the lines) and ODD dataset (everywhere else).}
    \label{fig:combined_model_comparison}
\end{figure}

This experiment demonstrates \ourmethod{}'s capability to handle uncertainties in a synthetic regression problem. The regression task is defined as follows:
\begin{equation}
\begin{aligned}
    f(x) &= \sin\left(\frac{x}{2}\right), \quad &x \in [-1, 1], \\
    \epsilon_i &\sim \mathcal{N}\left(0, \sigma(x)^2\right), \quad &\sigma(x)^2 = \frac{1 - |x|}{16}, \\
    y_i &= f(x_i) + \epsilon_i, 
\end{aligned} \label{eq:fwd-bay-reg}
\end{equation}
where \(y_i\) are noisy observations with state-dependent inhomogeneous noise. The goal is to learn the function \(f(x)\) while estimating two sources of uncertainty:
data(Aleatoric) uncertainty arising from state-dependent input noise in the data, and the model(epistemic) uncertainty due to the gap between the training data and inference points.  \Cref{fig:combined_model_comparison} shoes the dataset for this problem. 

The results are compared against various deep learning uncertainty quantification methods, including a Simple Neural Network with aleatoric uncertainty estimate (\texttt{SNN}), a Bayesian Neural Network (\texttt{BNN}), a Monte Carlo Dropout Network (\texttt{MCDO}), and a Deep Ensemble Network (\texttt{DENN}). Each architecture contains approximately 3,800--4,000 parameters to ensure a fair comparison. The data are split into a training set, an in-training-distribution (IDD) test set, and an out-of-training-distribution (OOD) test set. The latter division is used to gauge the predicted uncertainty of the model outside the training range: Ideally, the model would become less confident the further away the inference point is from the training data. In \cref{fig:combined_model_comparison}  the training range is denoted with vertical blue lines. 

\begin{highlightblue}

This test problem and the suite of uncertainty-aware deep neural networks are adopted from \cite{deepfindrColab2024}. For a comprehensive review of uncertainty quantification methods in deep learning, refer to \cite{abdar2021review}.  All networks are trained for 150 epochs with the \texttt{Adam} optimizer (batch size 16). Each model is run ten times with random initializations, and accuracies are averaged to mitigate random effects. Common hyperparameters are kept consistent across all models, while method-specific hyperparameters use default library values. To limit computational resource requirements, we have not performed method-specific hyperparameter tuning. 
    
\end{highlightblue}

We report training and testing mean squared errors (MSE) and the coverage of 95\% confidence intervals (CI) predicted by the networks. This design facilitates an evaluation of both predictive accuracy and the quality of uncertainty estimates within and outside the training range.

\begin{table}[htb]
    \centering
    \caption{Mean and standard deviation of testing MSE over 10 runs}
    \label{tab:compare_mean_std}
    \begin{tabular}{|c|c|c|}
        \hline
        \textbf{Architecture} & \textbf{MSE (mean)} & \textbf{MSE (std)} \\
        \hline
        \texttt{SNN}  & 0.147 & 0.007 \\
        \texttt{BNN}  & 0.352 & 0.008 \\
        \texttt{MCDO} & 0.198 & 0.034 \\
        \texttt{DENN} & 0.154 & 0.012 \\
        \ourmethod & 0.143 & 0.005 \\
        \hline
    \end{tabular}
\end{table}

\begin{table}[htb]
    \centering
    \caption{95\% confidence-interval coverage (averaged across 10 runs)}
    \label{tab:compare_confidence_coverage}
    \begin{tabular}{|c|c|c|c|}
        \hline
        \textbf{Architecture} & \textbf{Total Testing} & \textbf{IDD Coverage} & \textbf{OOD Coverage} \\
        \hline
        \texttt{SNN}  & 63.45 & 65.92 & 48.00 \\
        \texttt{BNN}  & 13.30 & 8.57  & 21.16 \\
        \texttt{MCDO} & 52.15 & 41.35 & 72.49 \\
        \texttt{DENN} & 59.70 & 36.21 & 88.16 \\
        \ourmethod & 89.05 & 85.42 & 97.00 \\
        \hline
    \end{tabular}
\end{table}

The results demonstrate that \ourmethod{} outperforms other architectures by achieving consistently lower mean squared errors (MSE) and significantly higher confidence-interval coverage. Notably, \ourmethod{} excels in out-of-training-distribution scenarios, highlighting its robustness and reliability in quantifying uncertainty across diverse testing conditions.

\begin{highlightblue}
    Figure \ref{fig:combined_model_comparison} presents the uncertainty quantification for \ourmethod{} without trunk (SNN), Monte Carlo Dropout Network, and regular \ourmethod{}. SNN effectively predicts the uncertainty in the data within the training range accurately; however, it estimates relatively low uncertainty for data outside the training range. This behavior is attributed to the lack of any proper mechanism to quantify the uncertainty in the network parameters. On the other hand, MCDO predicts a higher uncertainty outside of the training range. This results from the activation perturbations induced by the dropout. Yet, MCDO exhibits overconfident prediction on inference points within the training range. Compared to figure \ref{fig:snn}, which depicts \ourmethod{} without the trunk, the regular \ourmethod{} effectively captures meaningful uncertainty by perturbing the activation of the intermediate layer of the network. As a result, \ourmethod{} accurately captures both sources of uncertainty for both data within and outside the training range as indicated in figure \ref{fig:dpo}.

\end{highlightblue}

\subsection{One-dimensional function approximation}
\label{subsec:sin_cube}
This experiment evaluates the ability of \ourmethod{} to approximate a one-dimensional function while simultaneously estimating an unknown parameter and quantifying associated uncertainties. The function to be approximated is defined as:
\begin{equation}
y(x) = \sin^3(\omega x), \label{eq:sin_cube}
\end{equation}
where \(x \in [-1, 1]\) and \(\omega\) is an unknown parameter. For this experiment, the true value of \(\omega = 6\). The dataset consists of sampled values of \(x\) from the domain paired with noisy observations:
\[
y^{\text{data}}(x) = y(x) + \varepsilon, \quad \varepsilon \sim \mathcal{N}(0, \sigma). 
\]
Two noise scenarios are considered: a low-noise case with \(\sigma = 0.01\) and a high-noise case with \(\sigma = 0.1\).

\begin{figure}[h]
    \centering
    \subfloat[]{
        \includegraphics[height=90 pt]{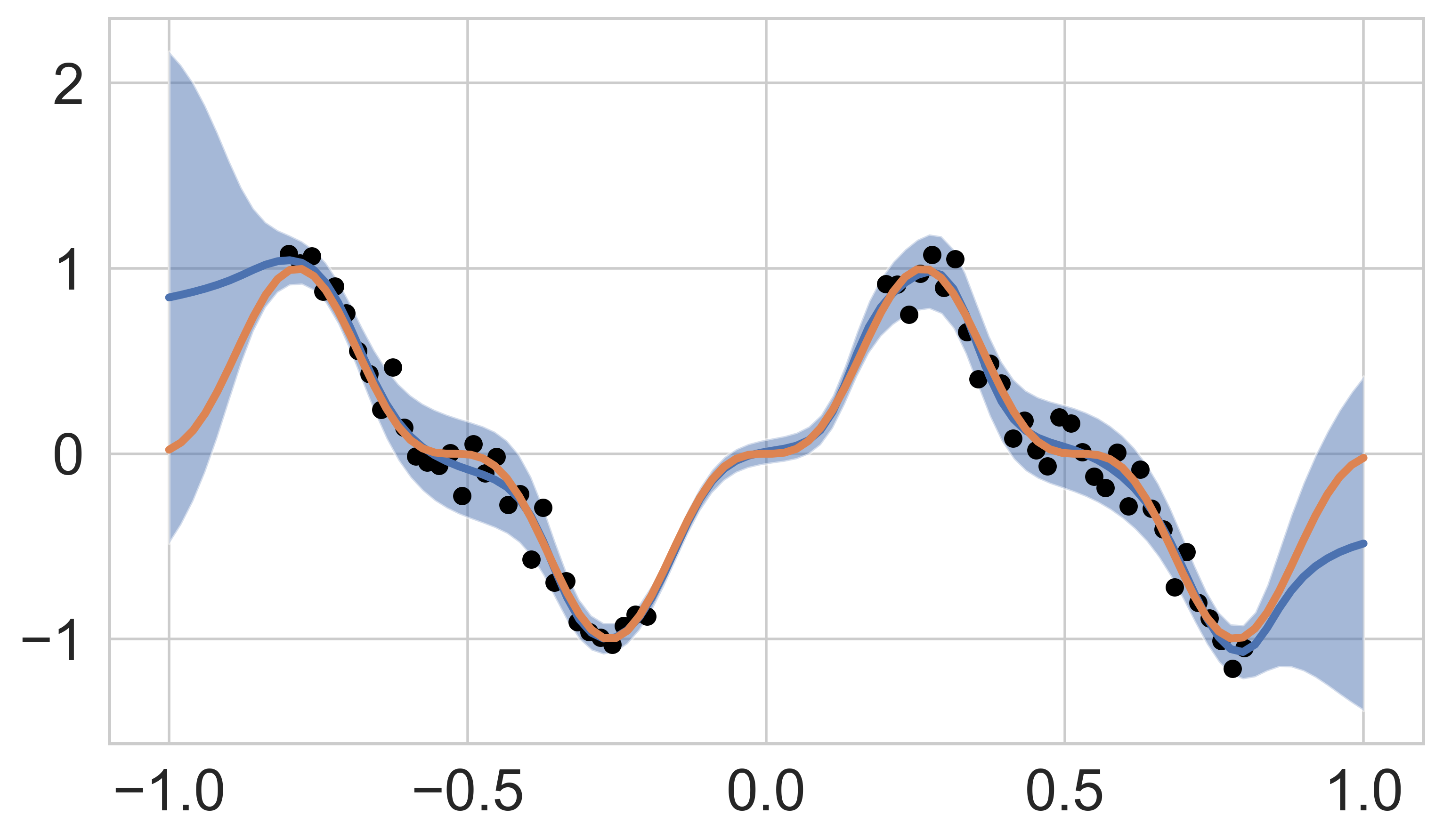}
    }
    \subfloat[ ]{
        \includegraphics[height=90 pt]{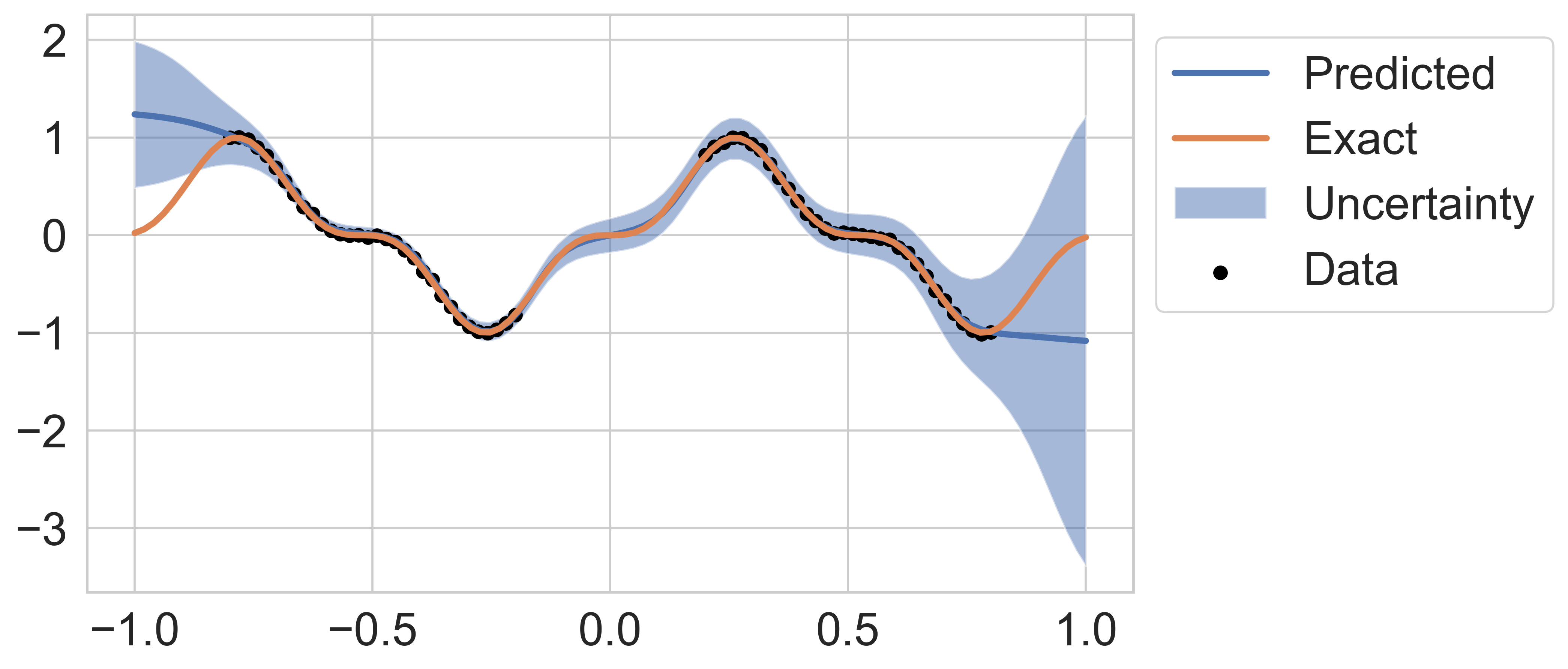}
    }\hfill
    \caption{\ourmethod learned solution for the regression problem in \cref{eq:sin_cube}. (a) $\sigma = 0.1$. (b) $\sigma = 0.01$}
    \label{fig:sin_uncertainity_combined}
\end{figure}

\Cref{fig:sin_uncertainity_combined} shows the mean predicted function as well as the two standard deviation uncertainty bands for both high- and low-noise cases. The model demonstrates good agreement with the data within the training range, while predicting elevated uncertainty outside the training range. Higher observation noise (\(\sigma = 0.1\)) results in more pronounced uncertainty, reflecting the model's ability to adapt to noisy data.

\begin{figure}[ht]
    \centering
    \subfloat[]{
        \includegraphics[width=0.45\linewidth]{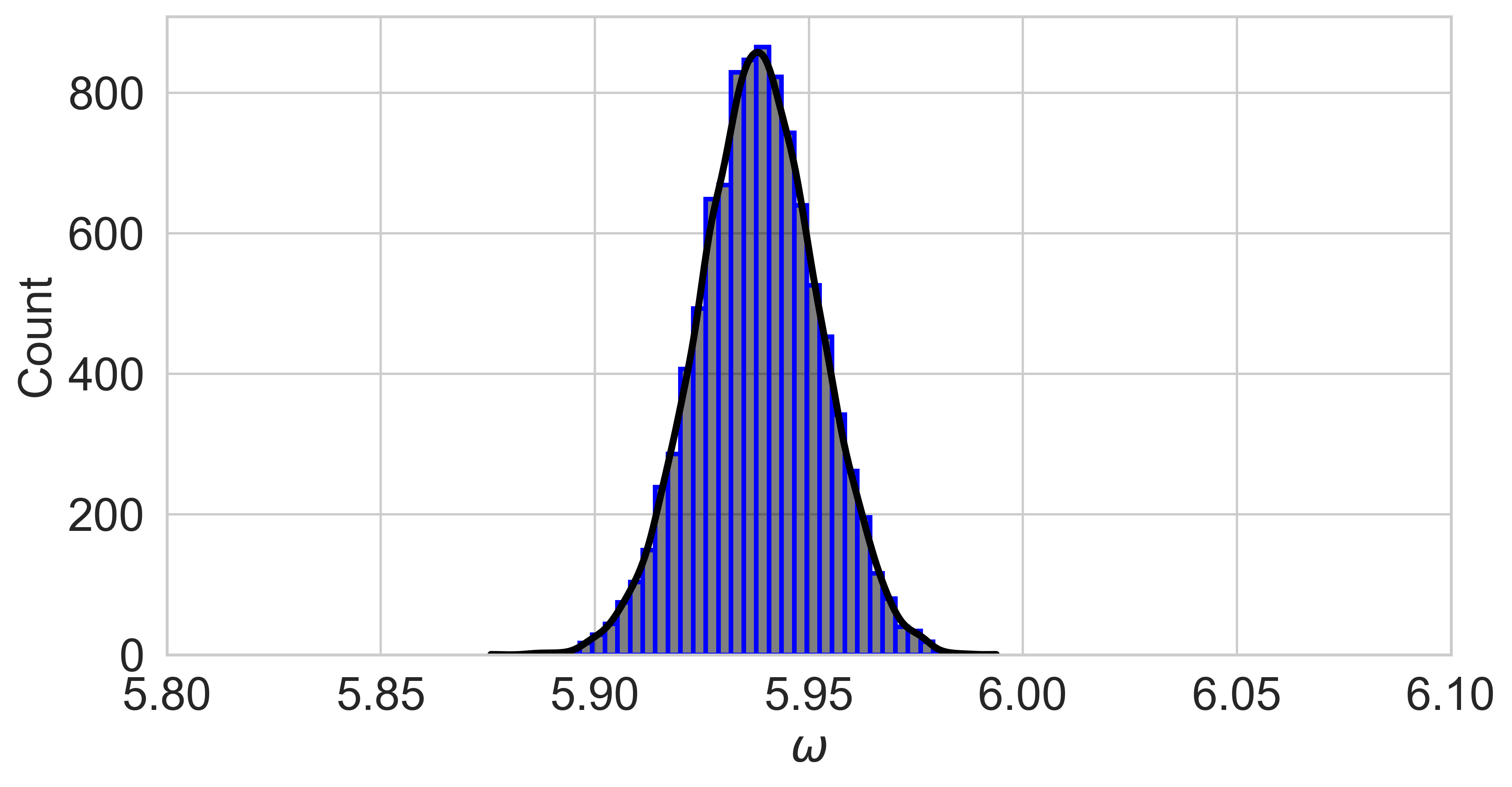}
    }
    \subfloat[ ]{
        \includegraphics[width=0.45\linewidth]{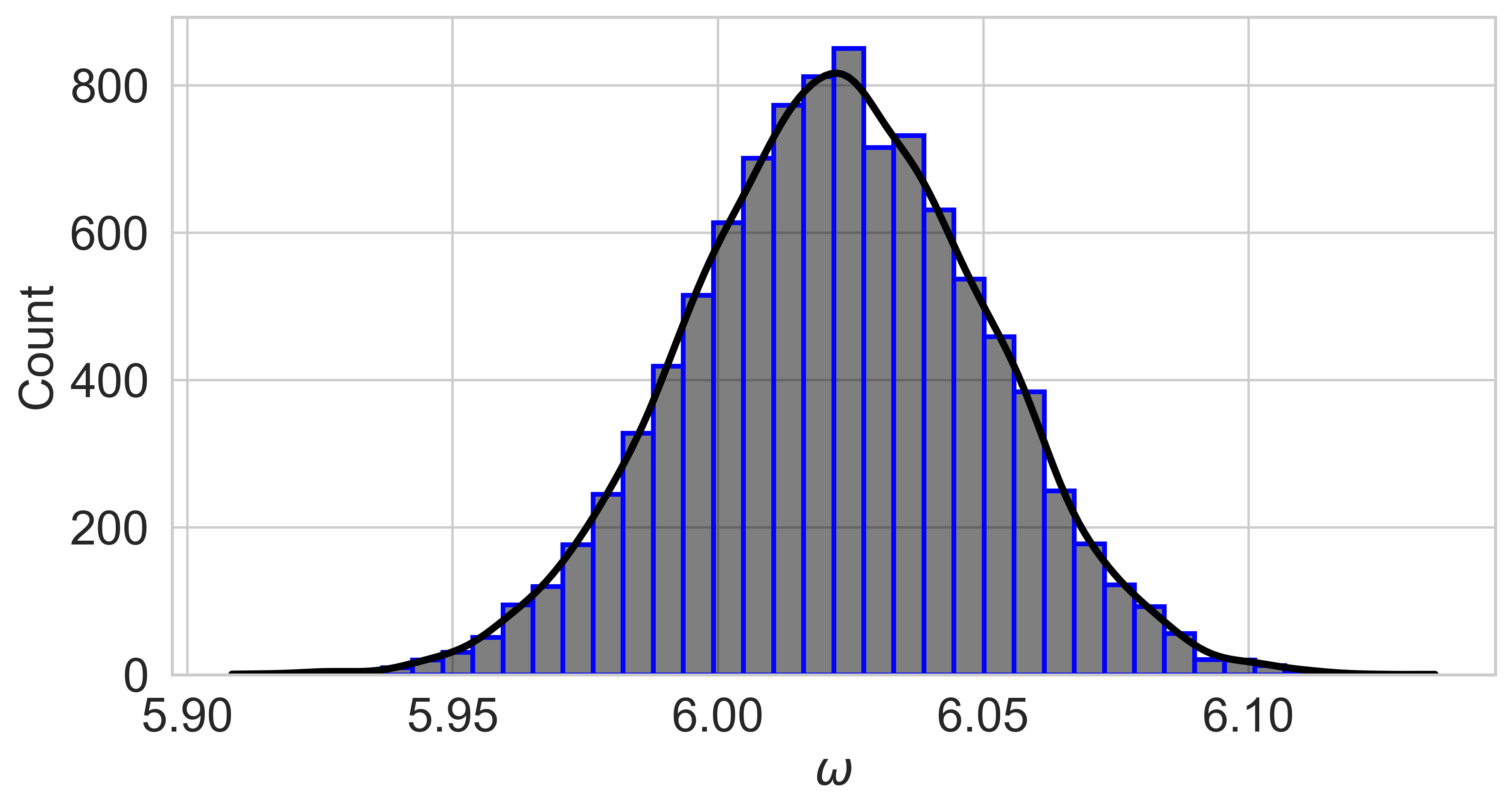}
    }\hfill
    \caption{Posterior of the parameter $\omega$ learned by \ourmethod from \cref{eq:sin_cube}.}
    \label{fig:sin_dist_combined}
\end{figure}

The posterior distribution of the parameter \(\omega\) is shown in \cref{fig:sin_dist_combined}, with the mode close to the true value of \(\omega = 6.0\) in both cases. The narrower posterior distribution in the low-noise scenario (\(\sigma = 0.01\)) indicates greater confidence in the parameter estimate compared to the high-noise scenario.

\subsection{One-dimensional unsteady heat equation}
\label{subsec:heat}

This experiment explores \ourmethod{}'s capability to solve a one-dimensional unsteady heat equation while simultaneously estimating the diffusion parameter \( D \) and the decay rate \( \alpha \). The heat equation is given as:
\begin{equation}
\pdv{y}{t} - D \pdv[2]{y}{x} = - e^{-\alpha t} \left( \sin(\pi x) - \pi^2 \sin(\pi x) \right), \quad x \in [-1, 1], \quad t \in [0, 1], \label{eq:heat}
\end{equation}
where \( D \) and \( \alpha \) are unknown parameters. When \( D = 1 \) and \( \alpha = 1 \), the equation has the exact solution:
\[
y(t, x) = e^{-t} \sin(\pi x).
\]

A \ourmethod{} model is trained to approximate this solution while learning the parameters \( D \) and \( \alpha \). The training dataset consists of 100 spatio-temporal samples with the corresponding exact solution \( y(t_i, x_i) \). A standard normal prior is used for both parameters. Training is performed for 15,000 epochs with a batch size of 100, using the \texttt{Adam} optimizer and an initial learning rate of 0.01. The loss component weights are configured as follows:
\[
w_\text{IC} = 3, \quad w_\text{Data} = 6, \quad w_\text{BC} = 1, \quad w_\text{Interior} = 1, \quad w_\text{STD} = 1.
\]


The model contains 2,103 trainable parameters.

\begin{highlightblue}

A basic grid search was employed to navigate the vast hyperparameter space and ensure the reproducibility of the experiments. Additionally, we implemented a learning rate scheduler that adaptively reduces the learning rate, further minimizing the model’s sensitivity to hyperparameters and improving its ability to navigate local minima. It’s worth noting that manually examining the magnitude of each component of the loss function \cref{eq:total_loss_practical} between training epochs is also an option to adjust them proportionally.    
\end{highlightblue}

\begin{figure}[h]
    \centering
    \subfloat[]{
        \includegraphics[width=0.31\textwidth]{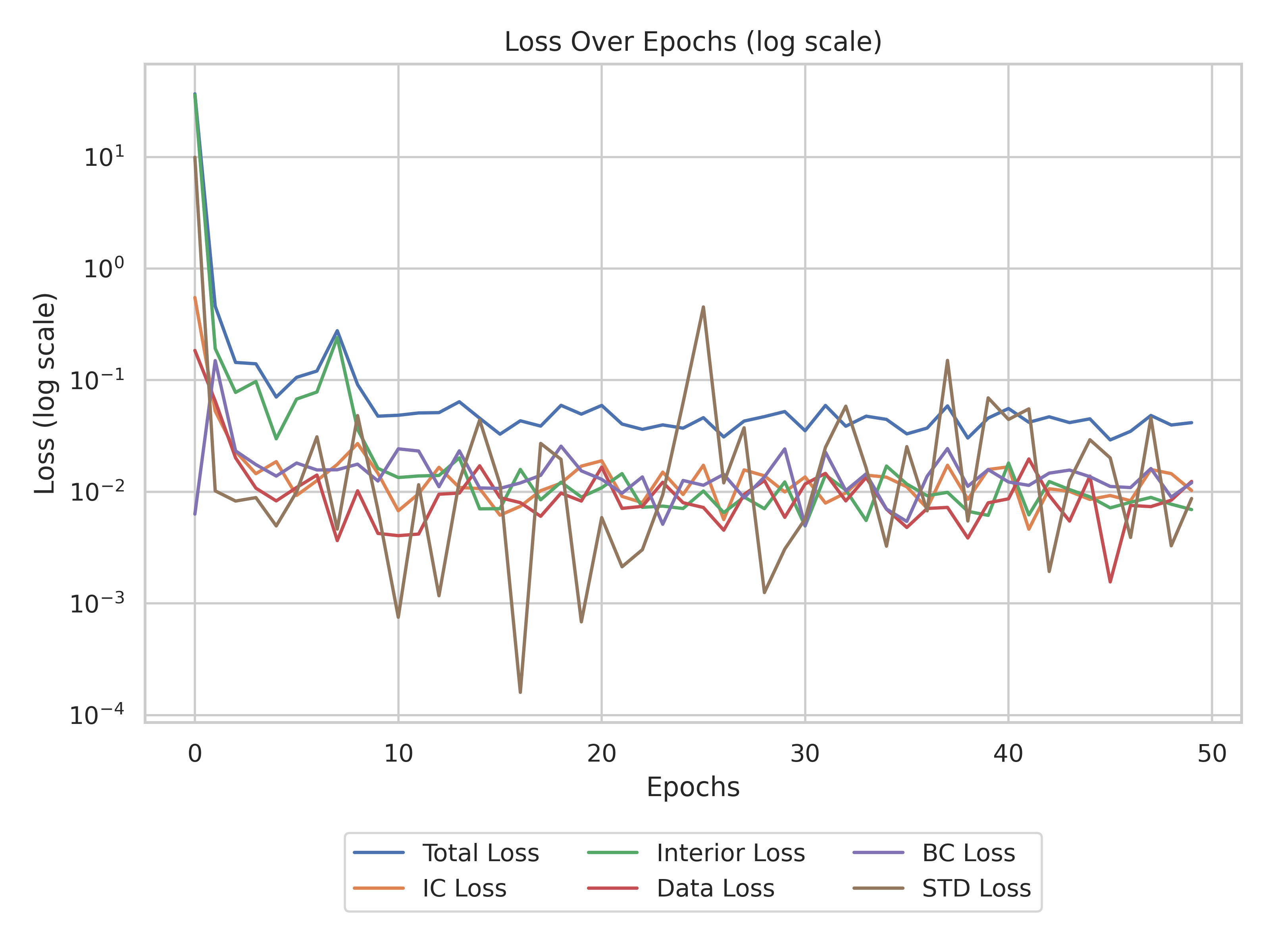}
        \label{fig:1D_log_loss}
    }
    \subfloat[]{
        \includegraphics[width=0.31\textwidth]{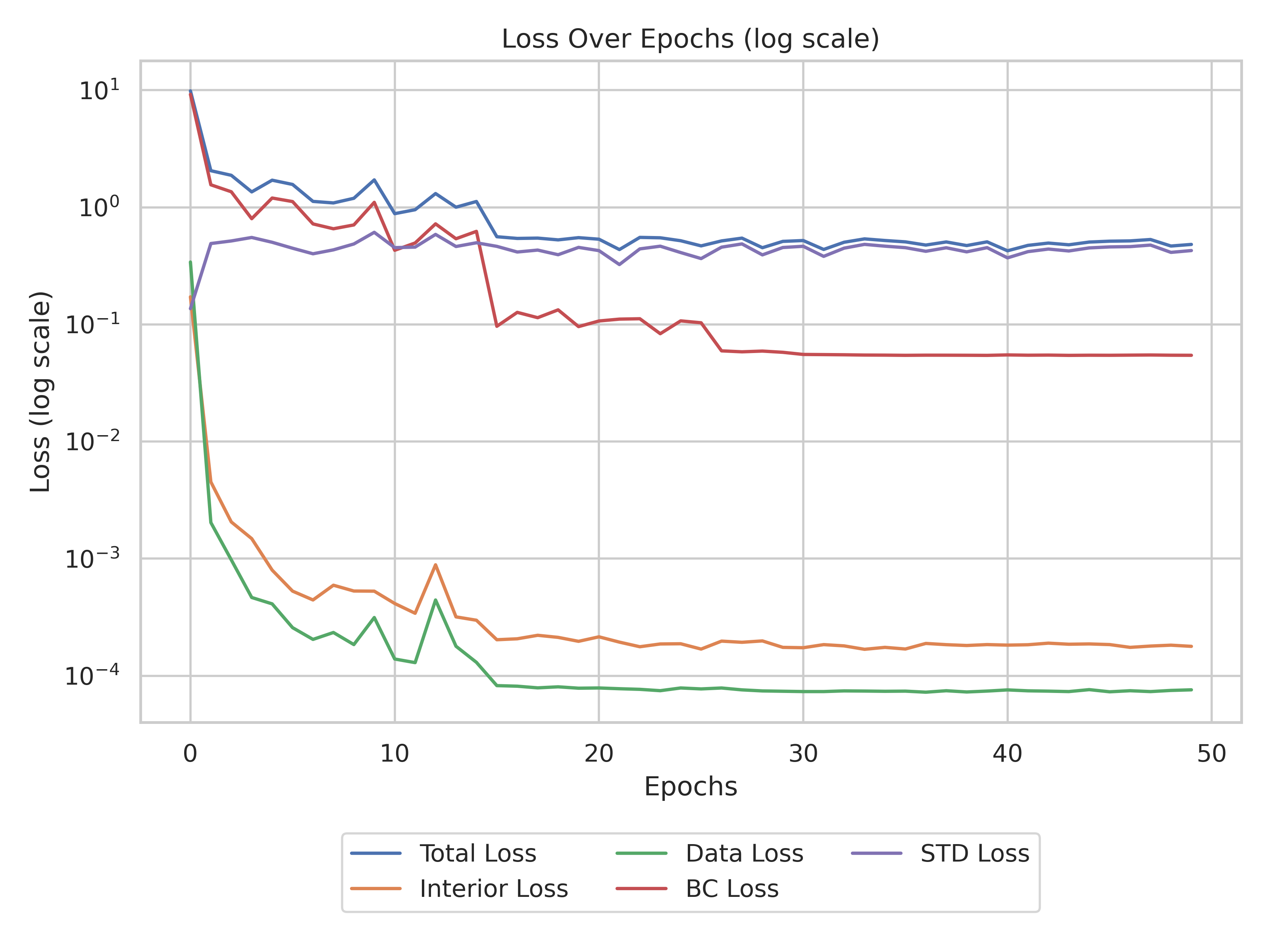}
        \label{fig:2D_log_loss}
    }
    \subfloat[]{
        \includegraphics[width=0.31\textwidth]{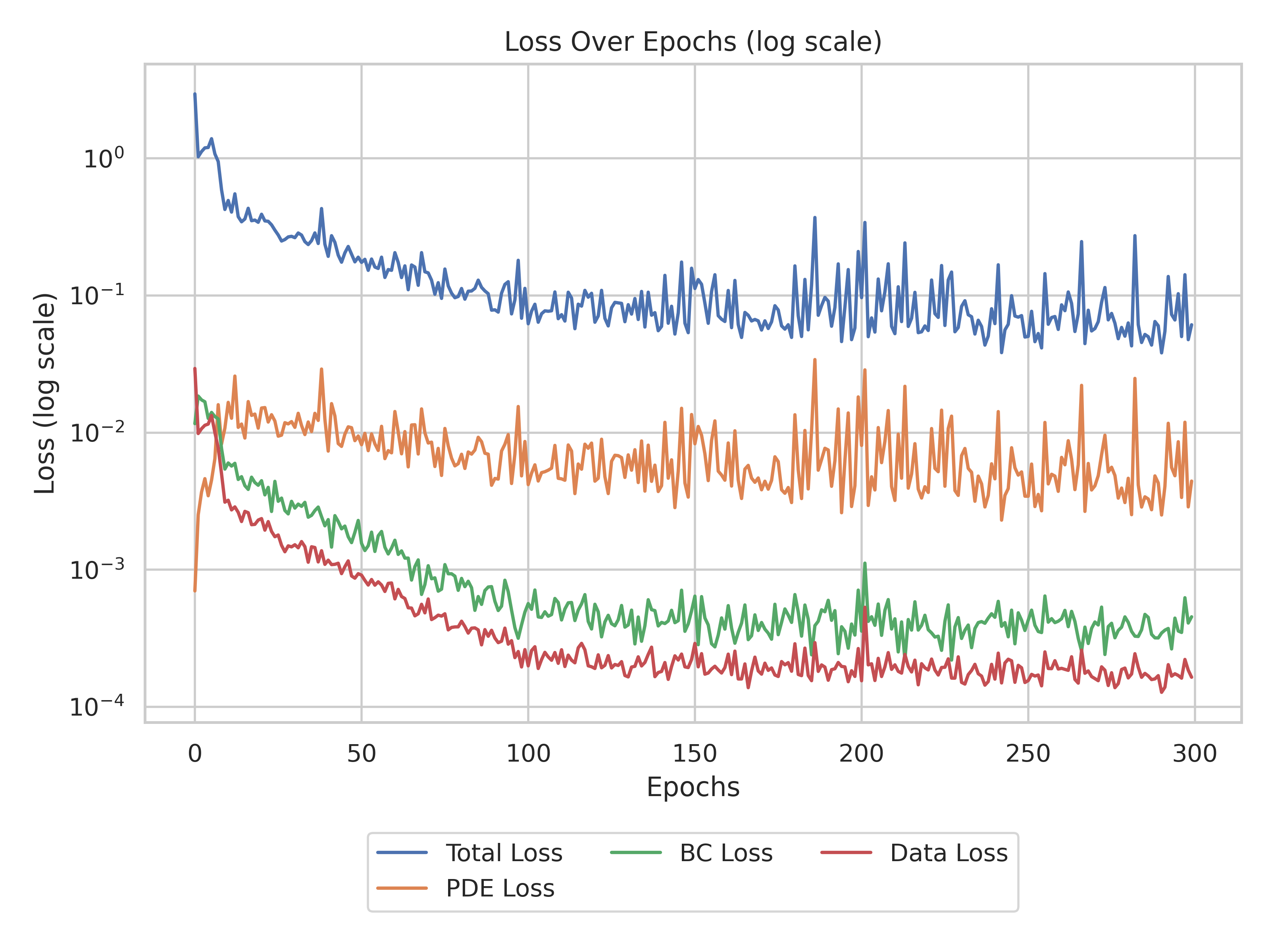}
        \label{fig:3D_log_loss}
    }
    \caption{(a) Loss history plot for 1D Heat equation in \cref{eq:heat} \cref{eq:heat}. (b) Loss history plot for the 2D Reaction-diffusion equation in \cref{eq:reaction-diffusion}. (c) Loss history plot for 3D Helmholtz equation in \cref{eq:3d-eig-problem} }
    \label{fig:combined_log_loss}
\end{figure}

During training, several metrics were recorded to monitor the model's performance and convergence, including the total loss and individual loss components: PDE (Interior) loss, boundary condition (BC) loss, initial condition (IC) loss, data loss, and the standard deviation loss for predicted parameters (STD loss). These metrics are shown in \cref{fig:1D_log_loss}, highlighting the stability and convergence of the model.

\begin{figure}[h]
    \centering
    \includegraphics[width=0.9\textwidth]{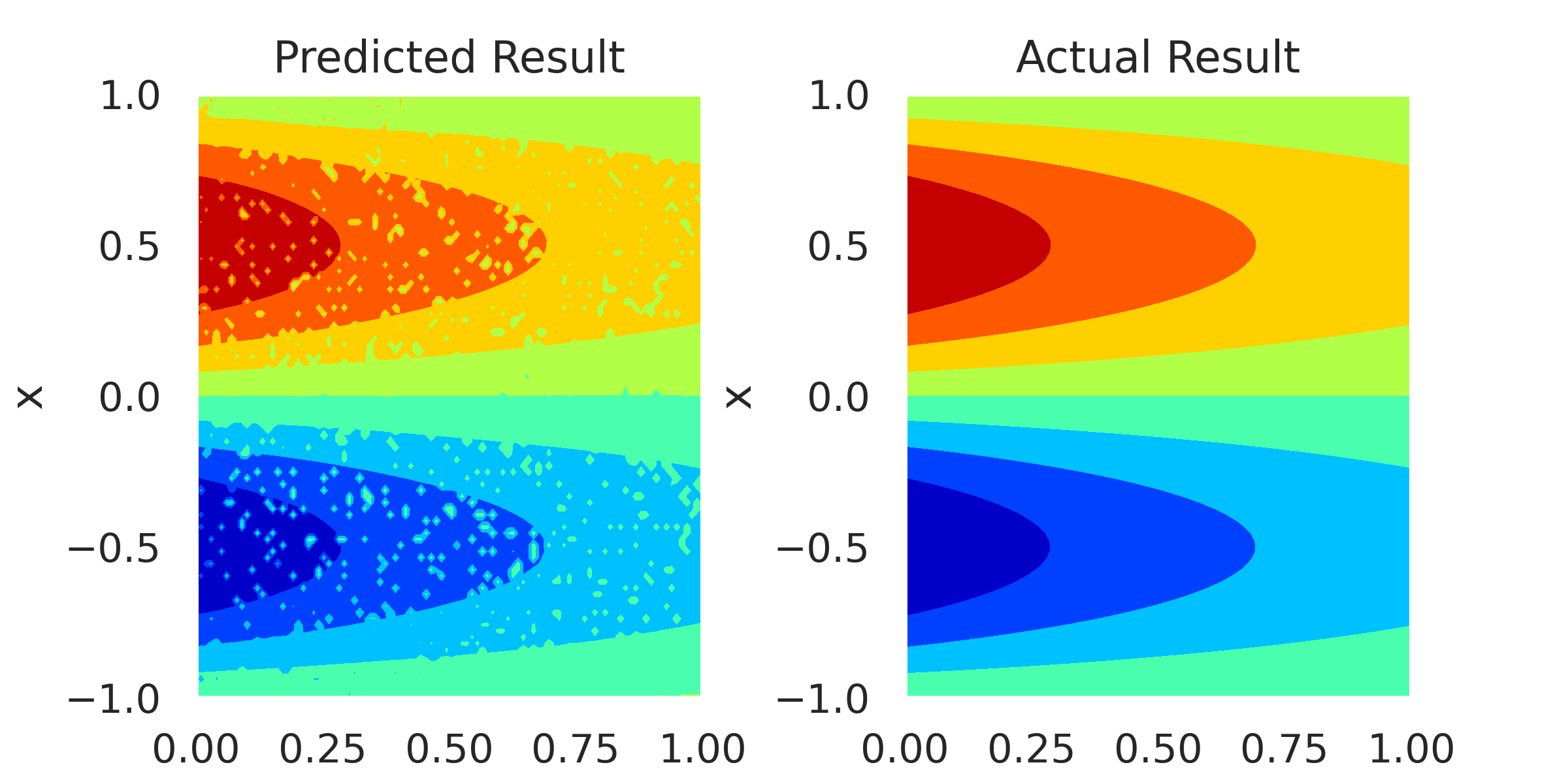}
    \caption{A sample of the predicted solution (left) and the exact solution (right) for the 1D Heat equation in \cref{eq:heat}.}
    \label{fig:heat1d-pred-exact}
\end{figure}

\Cref{fig:heat1d-pred-exact} compares a sample of the posterior predicted solution \( y(t, x) \) with the exact solution. The grainy appearance of the predicted solution reflects the uncertainty in the parameters \( D \) and \( \alpha \), which affects the point-wise forward solution. However, obtaining multiple samples of the output and computing statistical measures, such as the mean and mode, is computationally efficient once the model is trained.

\begin{figure}
    \centering
        \includegraphics[width= 0.85\textwidth]{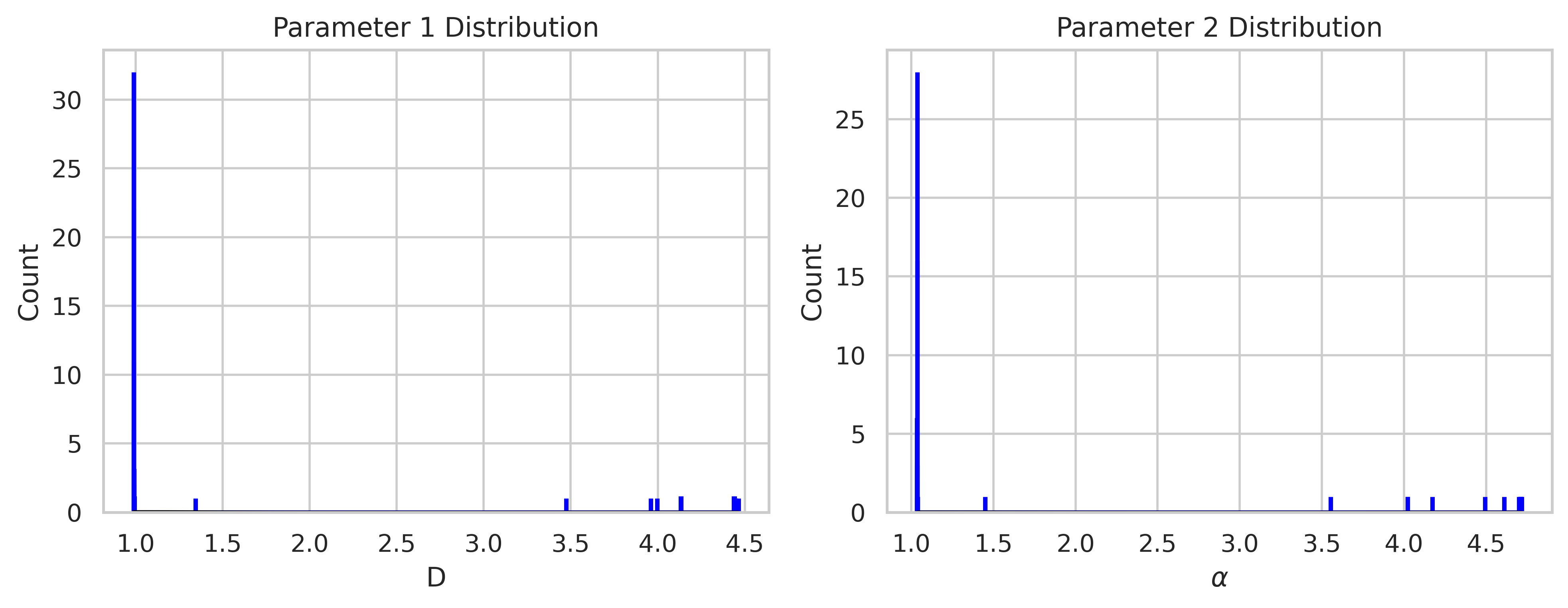}
        \caption{Histograms showing the distributions of Parameter \( D \) and \( \alpha \) for the 1D Heat equation in \cref{eq:heat}.}
        \label{fig:posterior_heat}
\end{figure}

The histograms in \cref{fig:posterior_heat} illustrate the posterior distributions of the parameters \( D \) and \( \alpha \). The model successfully predicts these parameters, with the posterior mode aligning closely with the true values (\( D = 1.0 \) and \( \alpha = 1.0 \)). Additionally, the secondary mode observed in the posterior distributions corresponds to larger values of \( D \) and \( \alpha \), indicating a highly diffusive and fast-decaying mode that still satisfies the PDE relatively well.

\begin{figure}[h]
    \centering
    \subfloat[]{
        \includegraphics[width=0.3\linewidth]{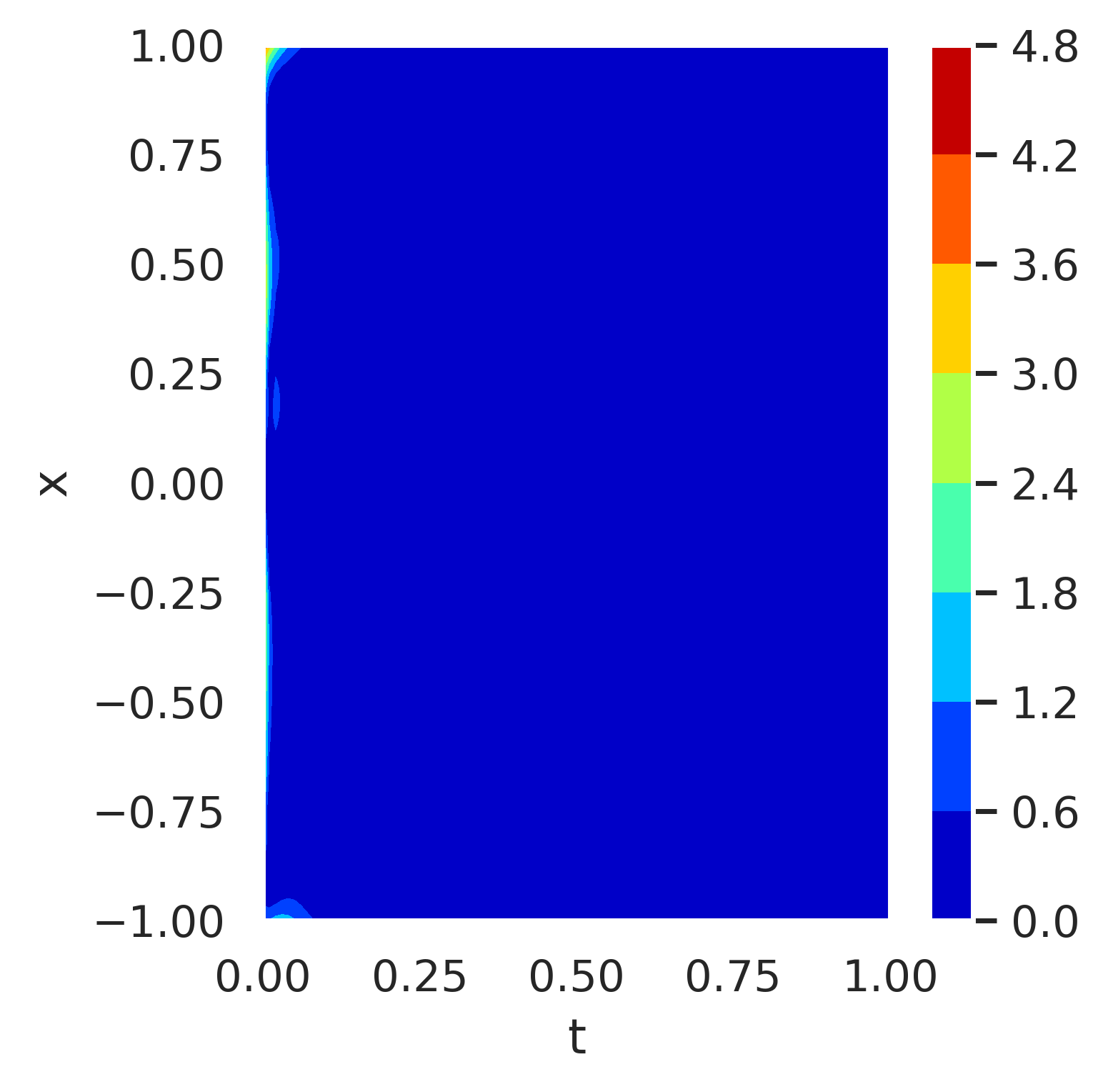} \label{fig:error_mean_1d_heat}
    }\hfill
    \subfloat[ ]{
        \includegraphics[width=0.3\linewidth]{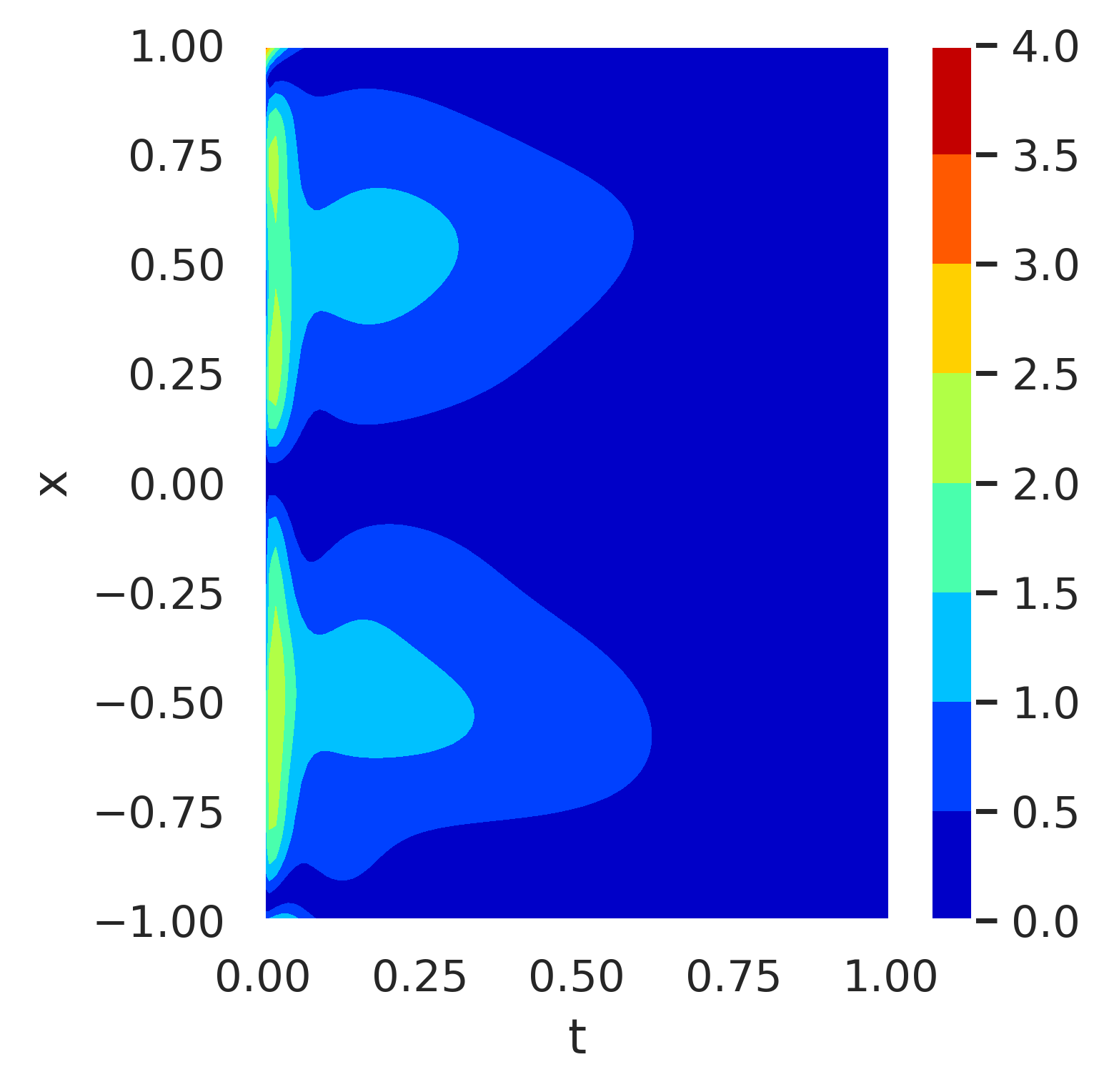} \label{fig:heat_residual_primary}}
    \hfill
    \subfloat[ ]{
        \includegraphics[width=0.3\linewidth]{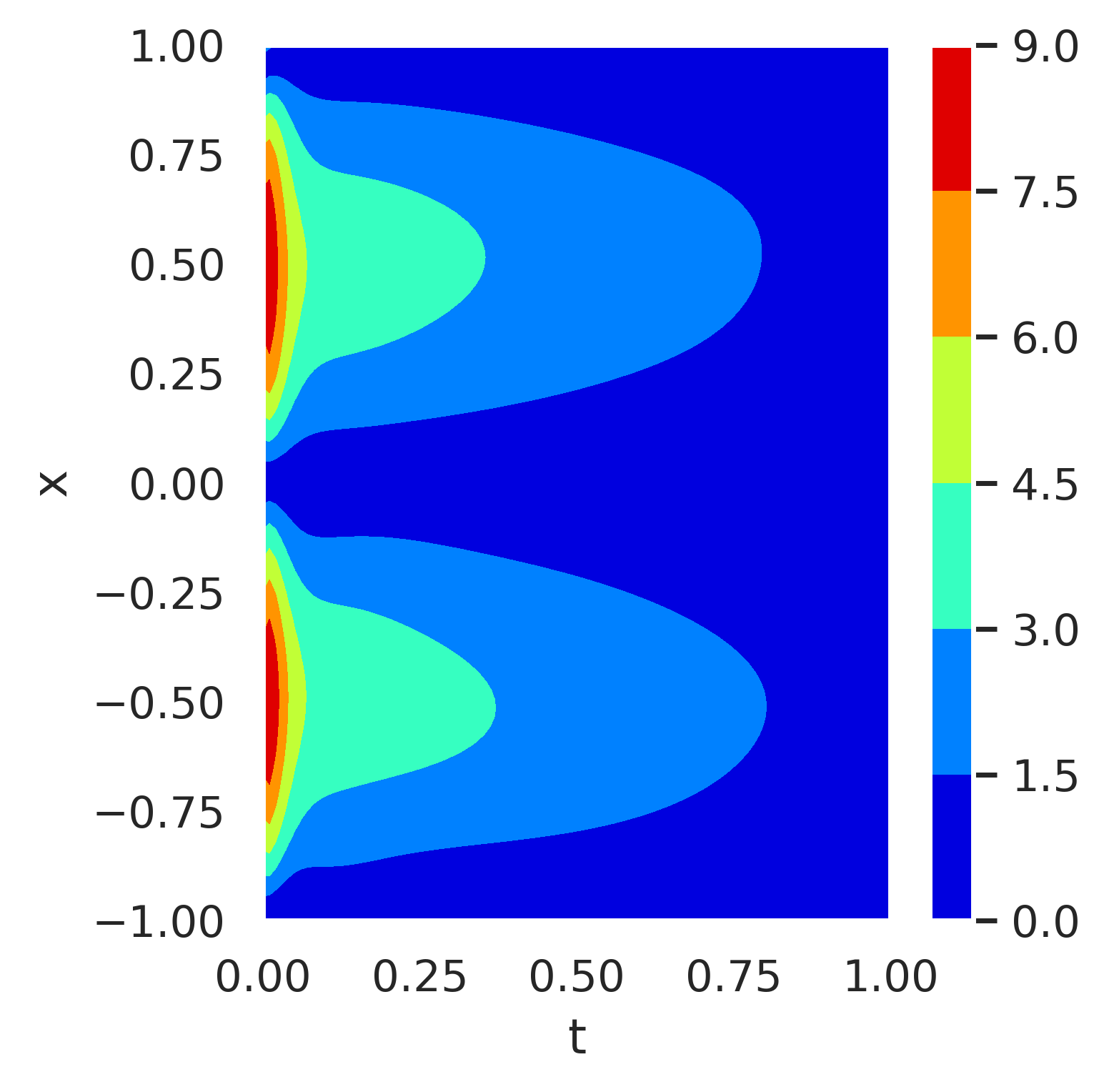} \label{fig:heat_residual_secondary}
    }
    \caption{Error and PDE residuals for the \ourmethod learned solution of the unsteady heat \cref{eq:heat}. (a) Forward solution absolute error compared to exact solution using the 
    mean of the parameter posterior distribution. (b) PDE residual using the primary mode of the parameter posterior distribution. (c) PDE residual using the secondary mode of the posterior parameter distribution.}
    \label{fig:heat_residual_combined}
\end{figure}

\Cref{fig:heat_residual_combined} further examines the errors and residuals for the \ourmethod{} learned solution. \Cref{fig:error_mean_1d_heat} shows the absolute error compared to the exact solution when using the mean of the posterior distribution of the parameters. Small errors are observed across the domain, except near the initial condition boundaries (\( t = 0, x = \pm 1 \)).

The PDE residuals, when evaluated using the primary mode (\cref{fig:heat_residual_primary}) and the secondary mode (\cref{fig:heat_residual_secondary}) of the posterior distribution, provide additional insights. The secondary mode results in significantly larger residuals, reflecting suboptimal parameter values that deviate from the true solution.

\subsection{Two-dimensional Reaction-diffusion equation}
\label{subsec:reaction-diffusion}

This experiment evaluates \ourmethod{}'s capability to solve a two-dimensional reaction-diffusion equation while estimating the reaction rate \( k \). The equation is defined as:
\begin{subequations}
\begin{align}
&\lambda \left( \frac{\partial^2 u}{\partial x^2} + \frac{\partial^2 u}{\partial y^2} \right) + k u^2 = f, \quad x, y \in [-1,1]^2, \\
&f(x, y) = \frac{1}{50} - \pi^2 u(x, y) + u(x, y)^2, \\
&u(x, y) = \sin(\pi x) \sin(\pi y). \label{eq:reaction-diffusion-solution}
\end{align}
\label{eq:reaction-diffusion}
\end{subequations}

For \( k = 1 \) and \( \lambda = 0.01 \), the PDE has a known solution as given in \cref{eq:reaction-diffusion-solution}. The training dataset consists of 100 randomly placed sensors within the domain that observe \( u \) and \( f \), with added observation noise of variance \( 0.01^2 \). Additionally, 25 sensors are uniformly placed along each boundary to enforce Dirichlet boundary conditions.

\begin{figure}[ht]
    \centering
    \subfloat[Training samples for \( u \)]{
        \includegraphics[width=0.45\linewidth]{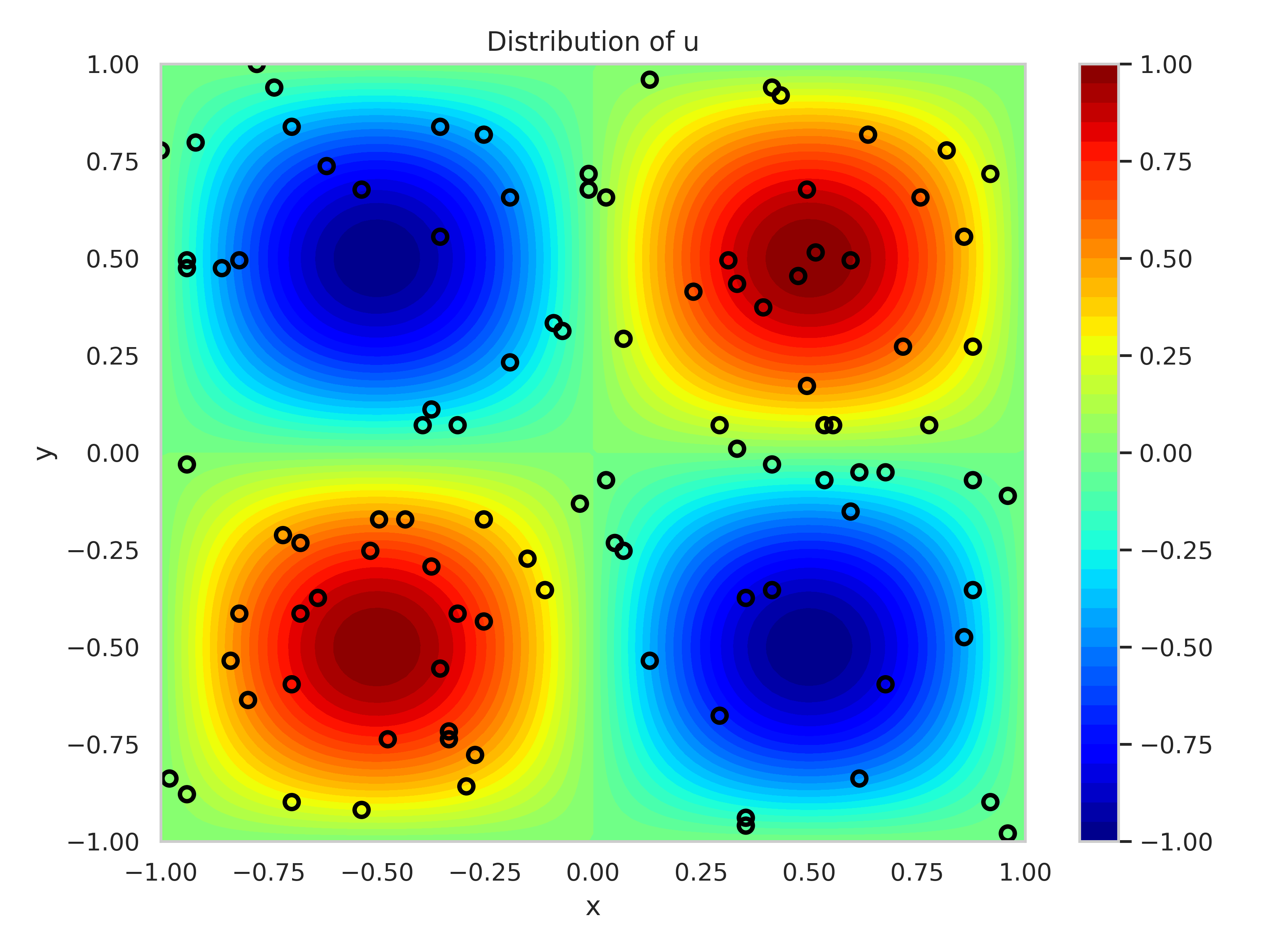}
    }
    \subfloat[Training samples for \( f \)]{
        \includegraphics[width=0.45\linewidth]{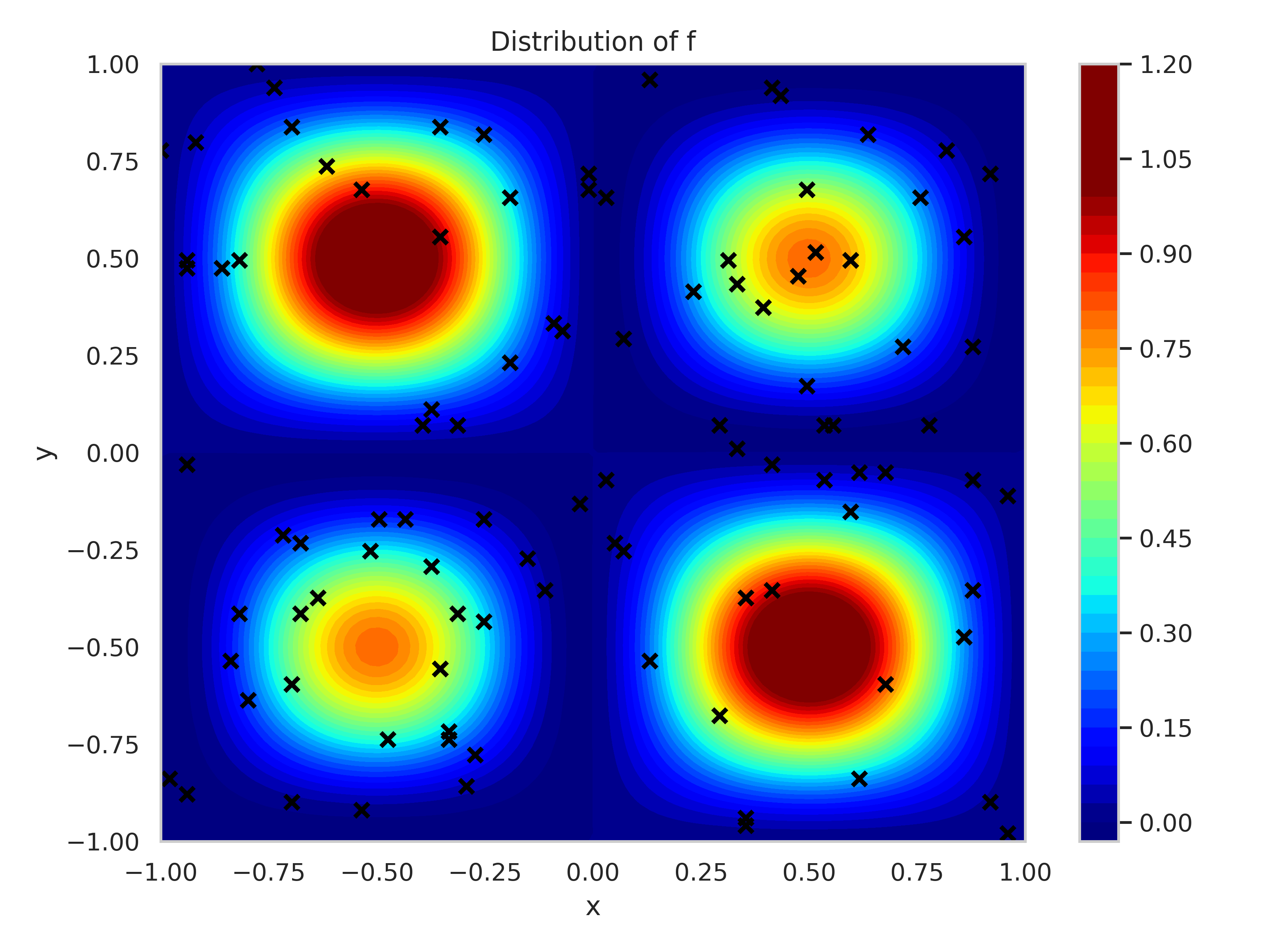}
    }\hfill
    \caption{Training samples, exact \( u \) and \( f \), and sensor positions for the 2D Reaction-diffusion equation in \cref{eq:reaction-diffusion}.}
    \label{fig:training_samples_2d}
\end{figure}

\Cref{fig:training_samples_2d} shows the sensor positions along with contours of \( u \) and \( f \). The model is configured with 300 neurons per hidden layer, totaling 272,703 trainable parameters. It is trained for 35,000 epochs using the \texttt{Adam} optimizer with a batch size of 500. Loss component weights are calibrated as:
\[
w_\text{Interior} = 20,000, \quad w_\text{Data} = 60,000, \quad w_\text{BC} = 100, \quad w_\text{STD} = 20.
\]
This setup ensures the optimizer appropriately balances the various constraints.

\begin{highlightblue}
Comparing the training loss history, as shown in \cref{fig:combined_log_loss}, we observe that the reaction-diffusion losses in \cref{fig:2D_log_loss} have magnitudes of difference, while in \cref{fig:1D_log_loss} the loss components are well-balanced. This is due to the fact that in \cref{fig:1D_log_loss} the weights for different losses are about the same while the data loss and interior loss for the reaction-diffusion equation had to be scaled up dramatically for a successful training.
\end{highlightblue}

\Cref{tab:comparision_2d_reaction} compares the estimated mean and standard deviation of the reaction rate \( k \) using \ourmethod{} and alternative Bayesian methods. \ourmethod{} achieves results closest to the true value \( k = 1 \) with the lowest standard deviation, demonstrating both accuracy and certainty in parameter estimation.

\begin{table}[ht]
    \centering
    \resizebox{0.85 \linewidth}{!}{%
    \begin{tabular}{lccccc}
    \hline
    \textbf{} & {\ourmethod} & \texttt{B-PINN-HMC} & \texttt{B-PINN-VI} & \texttt{Dropout-1\%} & \texttt{Dropout-5\%} \\
    \hline
    \textbf{Mean} & 0.999 & 1.003 & 0.895 & 1.050 & 1.168 \\
    \textbf{Std} & $5.4 \times 10^{-4}$ & $5.75 \times 10^{-3}$ & $2.83 \times 10^{-3}$ & $2.00 \times 10^{-3}$ & $3.04 \times 10^{-3}$ \\
    \hline
    \end{tabular}
    } 
    \caption{2D nonlinear diffusion-reaction system: Predicted mean and standard deviation for the reaction rate \( k \) using different Bayesian PINN methods. The exact solution is \( k = 1 \). Table reproduced from \cite{bpinns2021yang} with added data.}
    \label{tab:comparision_2d_reaction}
\end{table}

\begin{figure}[ht]
    \centering
    \subfloat[With boundary conditions]{
        \includegraphics[width=0.44\linewidth]{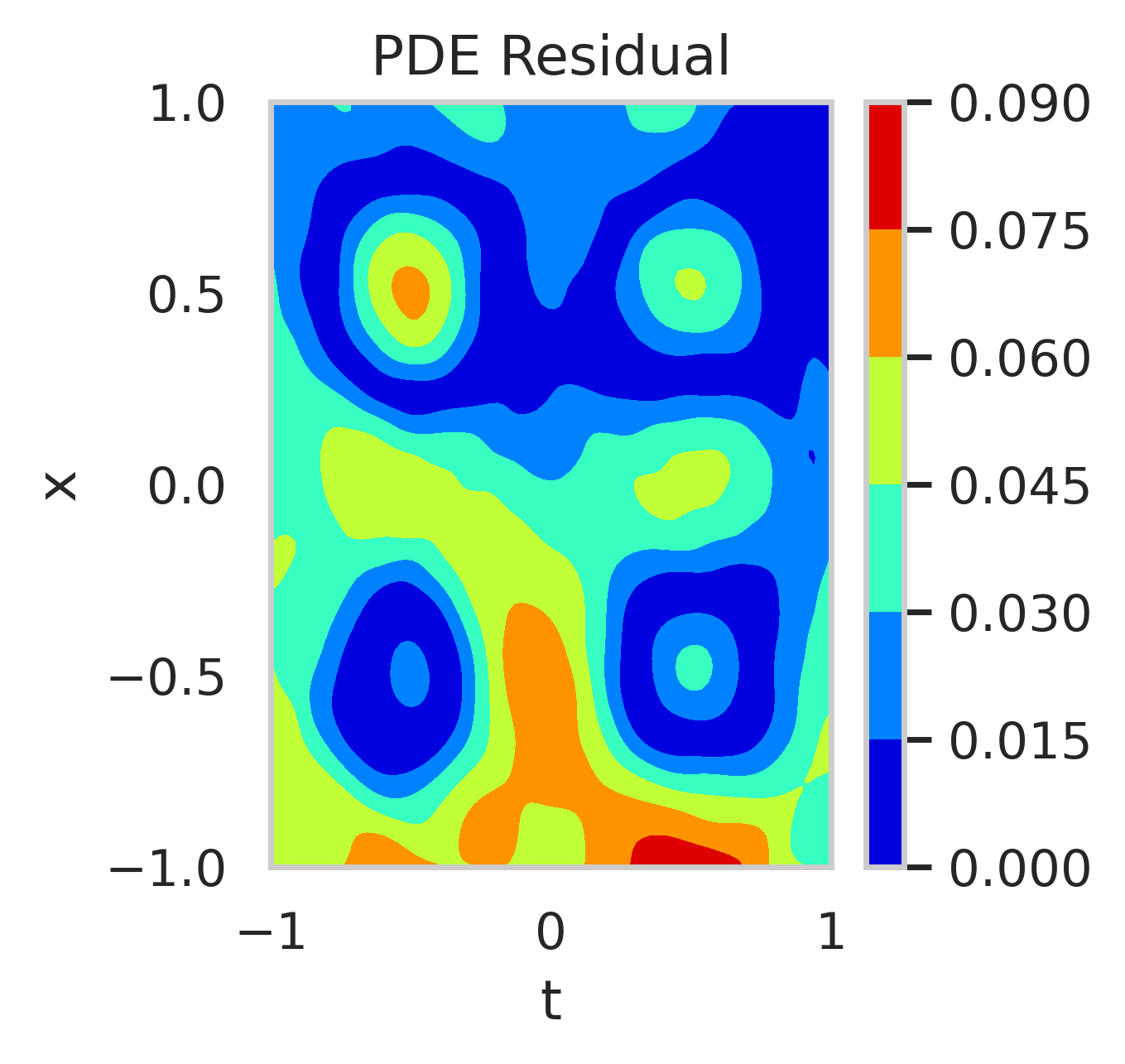}
    }
    \subfloat[ Without boundary conditions]{
        \includegraphics[width=0.44\linewidth]{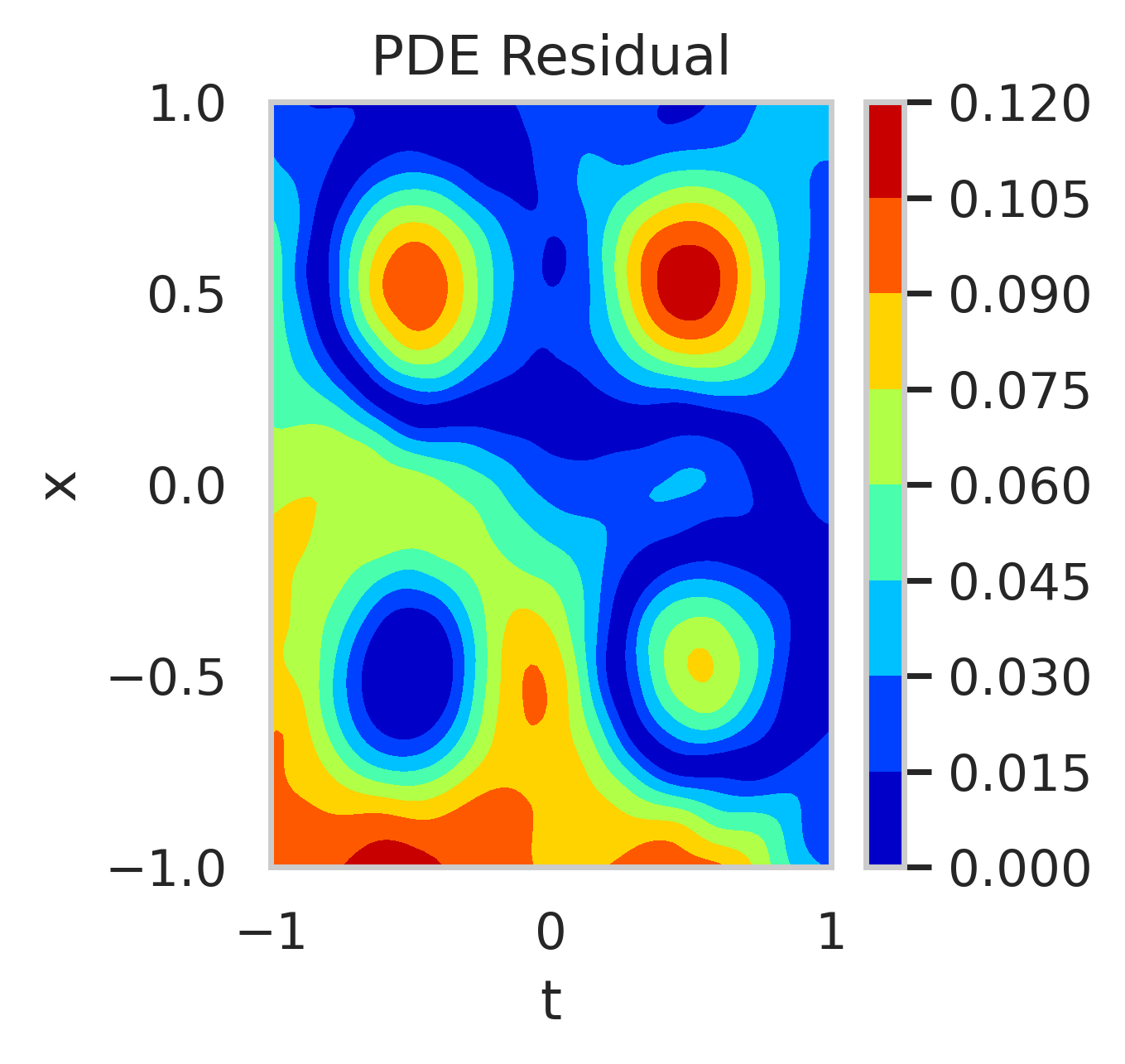}
    }\hfill
    \caption{PDE residuals for the 2D Reaction-diffusion equation in \cref{eq:reaction-diffusion}.}
    \label{fig:combined_residual_plots}
\end{figure}

We conducted a second experiment by relaxing boundary condition enforcement (\( w_\text{BC} = 0 \)). \Cref{fig:combined_residual_plots} shows the residuals of the PDE when using the mean posterior parameter value. Higher residuals are observed at the boundaries when boundary conditions are excluded from training.

\begin{figure}[ht]
    \centering
    \subfloat[With boundary conditions]{
        \includegraphics[width=0.45\linewidth]{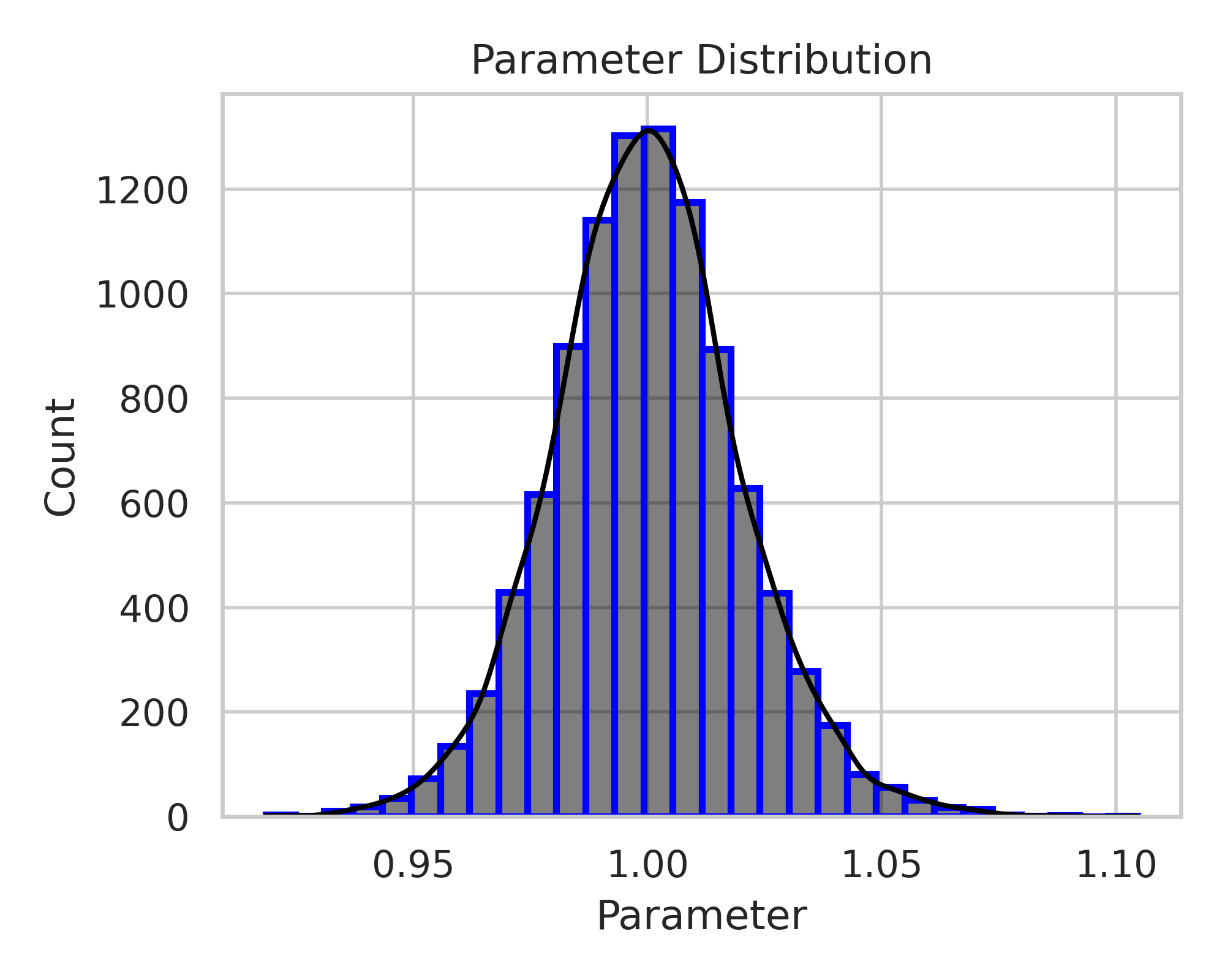}
    }\hfill
    \subfloat[Without boundary conditions]{
        \includegraphics[width=0.45\linewidth]{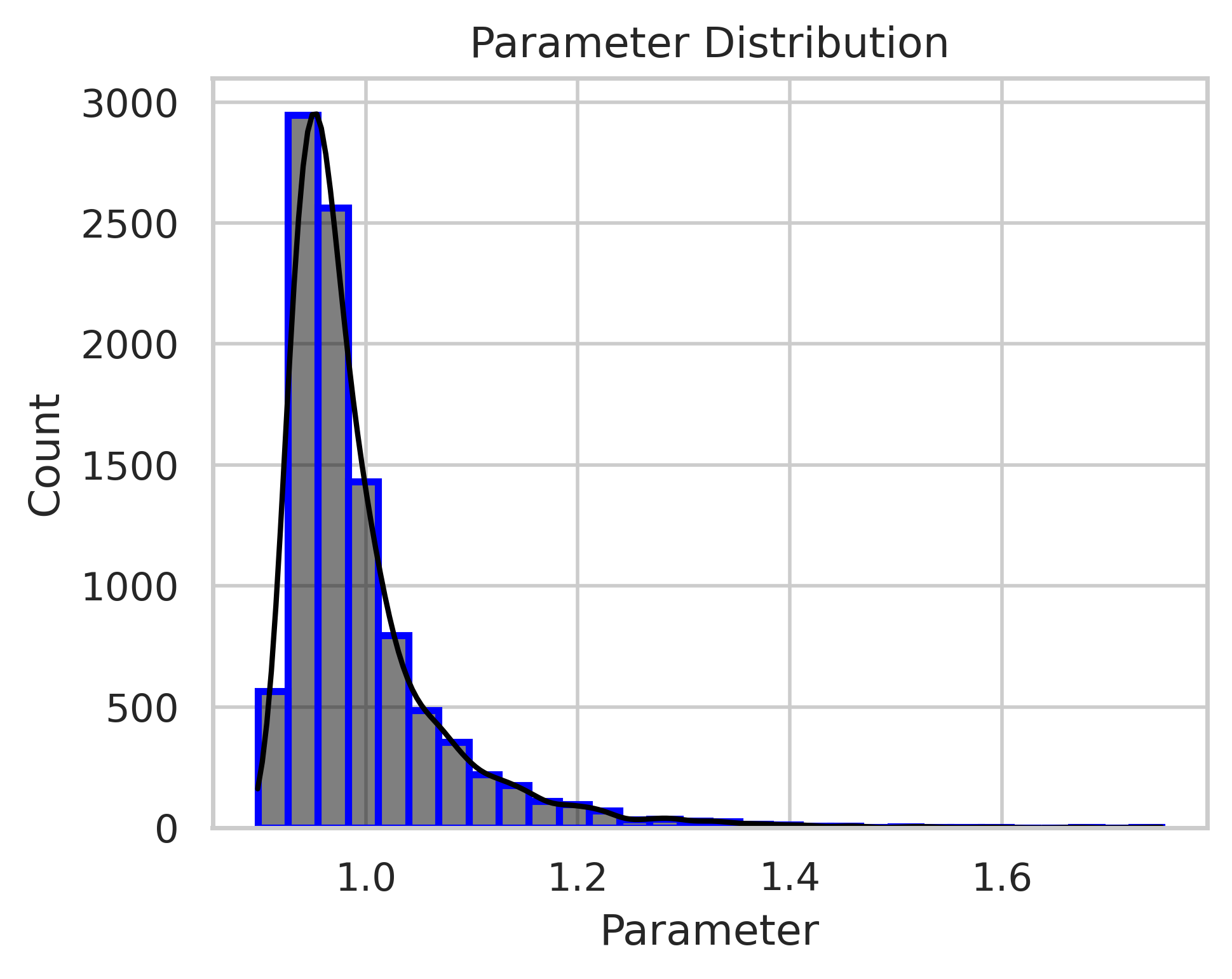}
    }
    \caption{Posterior distribution of the parameter \( k \) for the 2D Reaction-diffusion equation in \cref{eq:reaction-diffusion}.}
    \label{fig:2d_distribution_combined}
\end{figure}

\begin{figure}[ht]
    \centering
        \includegraphics[width=0.9\linewidth]{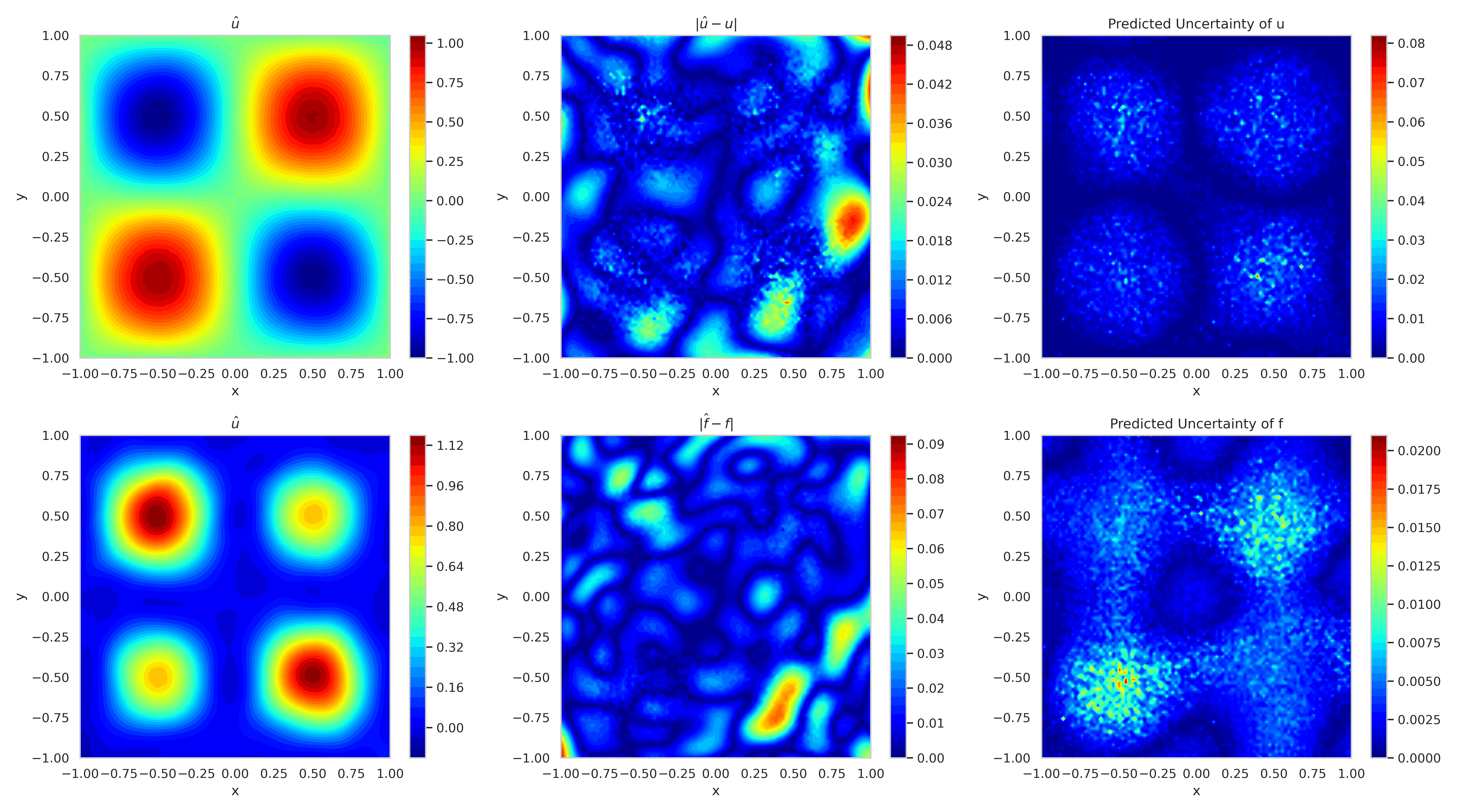}
    \caption{Predicted mean solution, absolute error, and uncertainty for the 2D Reaction-diffusion equation  in \cref{eq:reaction-diffusion} with boundary condition training.}
    \label{fig:2d_uncertainity_no_bound}
\end{figure}

The posterior distributions of \( k \) are shown in \cref{fig:2d_distribution_combined}. With boundary conditions, the posterior is narrow and centered around \( k = 1.0 \). Without boundary conditions, the posterior is wider, skewed, and exhibits a long tail, indicating increased uncertainty towards positive reaction rates.
\begin{figure}[ht]
    \centering
        \includegraphics[width=0.9\linewidth]{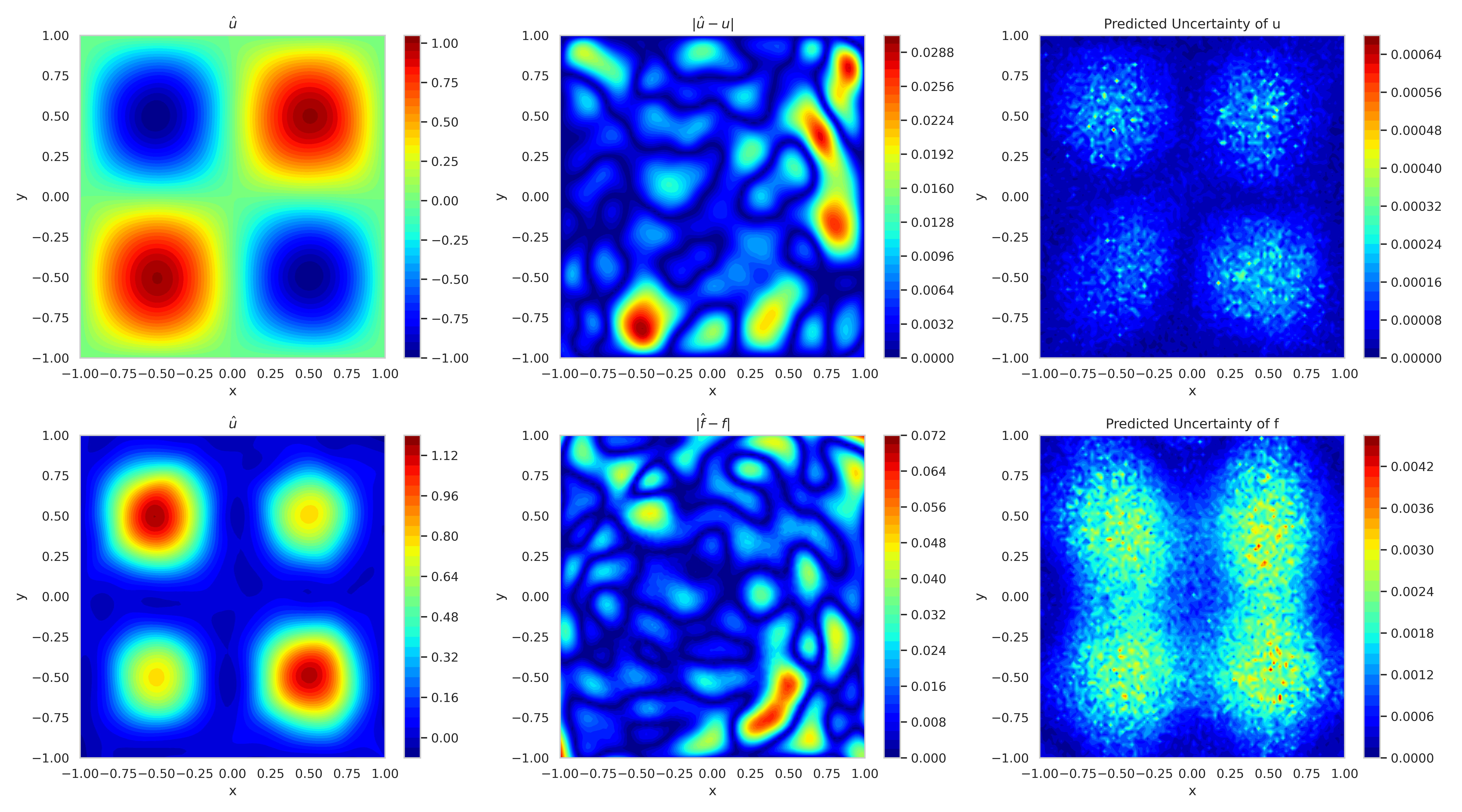}
    \caption{Predicted mean solution, absolute error, and uncertainty for the 2D Reaction-diffusion equation  in \cref{eq:reaction-diffusion} without boundary condition training.}
    \label{fig:2d_uncertainity}
\end{figure}
\Cref{fig:2d_uncertainity} shows the mean predicted solution and its error along with the predicted uncertainty for boundary-enforced training. In \cref{fig:2d_uncertainity_no_bound}, the corresponding plots for boundary-free training reveal reduced accuracy and elevated uncertainty, highlighting the ability of the model to adapt to missing boundary conditions.

\subsection{3D Eigenvalue problem in a unit ball}
\label{subsec:3d-eig}

In this experiment, we use the three--dimensional Dirichlet--Laplacian 3D problem:
\begin{equation}
\begin{cases}
-\Delta u(\mathbf x)=\lambda\,u(\mathbf x), & \|\mathbf x\|<1,\\[4pt]
u(\mathbf x)=0, & \|\mathbf x\|=1.
\end{cases}
\label{eq:3d-eig-problem}
\end{equation}

In Cartesian coordinates the domain boundary
$x^{2}+y^{2}+z^{2}=1$ cuts through a regular grid at oblique angles.
 Conventional grid-based solvers must either deform a Cartesian mesh to fit the curved boundary or use immersed-boundary tricks. Both options complicate stencil design. Furthermore, estimating the eigenvalue~$\lambda$ as a parameter depends on the entire error in~$u$,
not just local residuals, which makes accurate estimation of both
quantities challenging.

We solve this problem using \ourmethod{} by approximating the eigenmode via the branch network and the eigenvalue using the trunk network. 
A numerical challenge in learning the eigenpairs is that the PDE and the boundary conditions are satisfied by
$u\equiv0$ for any~$\lambda$, so additional constraints
are required to
force the network toward a non‐trivial eigenmode.

\[
\mathcal L=
\underbrace{\mathbb E_{\Omega}\!\bigl[-\Delta u_{\theta}-\lambda_{\theta}u_{\theta}\bigr]^{2}}_{\text{PDE residual}}
\;+\;
\underbrace{\mathbb E_{\partial\Omega}\!\bigl[u_{\theta}^{2}\bigr]}_{\text{Dirichlet BC}}
\;+\;
\underbrace{\bigl(\mathbb E_{\Omega}[u_{\theta}^{2}]-1\bigr)^{2}}_{\text{normalization}}.
\]

The loss function for this problem combines three terms:
(i) the PDE residual $(-\Delta u-\lambda u)^{2}$ evaluated at random
interior points,
(ii) the squared boundary condition on the sphere,
and (iii) a simple normalization $(\mathbb E[u^{2}]-1)^{2}$ to guide the learned solution away from the trivial solution $u\equiv0$.
Automatic differentiation supplies exact gradients of all three terms,
so the optimizer can update the network
weights \emph{and} the spectral parameter~$\lambda$
in a single pass.

The eigen-functions of \cref{eq:3d-eig-problem} are available in closed form. They separate into spherical Bessel functions $j_\ell(k_{\ell s} r)$
and spherical harmonics $Y_\ell^m(\theta,\varphi)$. Our neural solver
never sees this analytic structure.  Throughout the experiments, the
network receives only Cartesian coordinates $(x,y,z)$ and must satisfy
the boundary condition $x^{2}+y^{2}+z^{2}=1$ implicitly through the
loss.  Working in Cartesian space, therefore, removes the coordinate
system in which the PDE decouples, forces the optimizer to discover a
non-trivial eigenfunction from scratch, and provides a more demanding
test of the physics-informed loss. The first two eigenvalues are
$\lambda_{01}=\pi^{2}\approx9.87$ (radial mode) and 
$\lambda_{11}=4.49341^{2}\approx20.19$ (dipole mode).
During training, we monitor convergence toward
$\lambda_{11}$.

\begin{figure}[ht]
    \centering

    \begin{minipage}{0.01\linewidth}
    \end{minipage}%
    \begin{minipage}{0.29\linewidth}
        \centering
        \scriptsize \hspace{-2mm}Predicted
    \end{minipage}%
    \begin{minipage}{0.29\linewidth}
        \centering
        \scriptsize \hspace{-5mm}Exact
    \end{minipage}%
    \begin{minipage}{0.29\linewidth}
        \centering
        \scriptsize \hspace{-7mm}Error
    \end{minipage}

    \vspace{4mm}
    \begin{minipage}{0.01\linewidth}
        \centering
        \rotatebox{90}{\scriptsize $z=0.00$}
    \end{minipage}%
    \begin{minipage}{0.93\linewidth}
        \centering
        \includegraphics[width=0.9\linewidth]{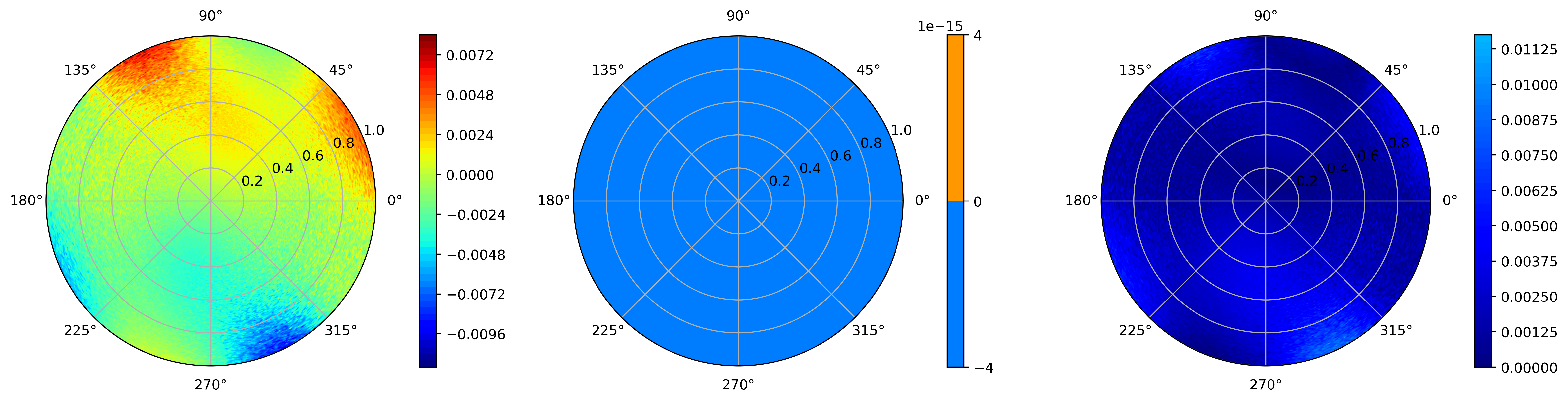}
    \end{minipage}

    \begin{minipage}{0.01\linewidth}
        \centering
        \rotatebox{90}{\scriptsize $z=0.25$}
    \end{minipage}%
    \begin{minipage}{0.93\linewidth}
        \centering
        \includegraphics[width=0.9\linewidth]{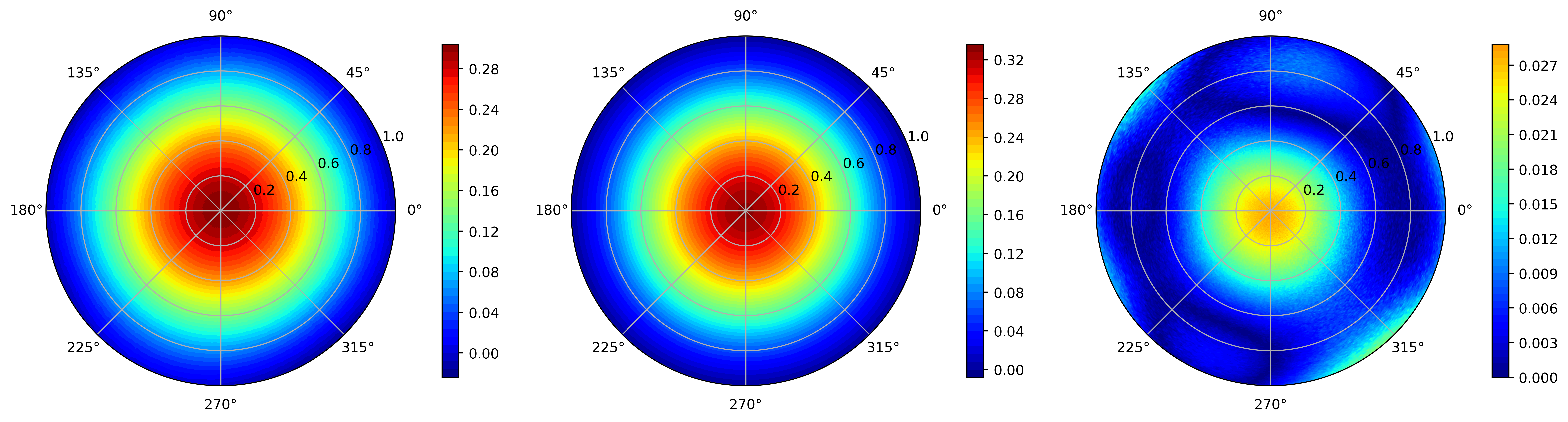}
    \end{minipage}

    \begin{minipage}{0.01\linewidth}
        \centering
        \rotatebox{90}{\scriptsize $z=0.50$}
    \end{minipage}%
    \begin{minipage}{0.93\linewidth}
        \centering
        \includegraphics[width=0.9\linewidth]{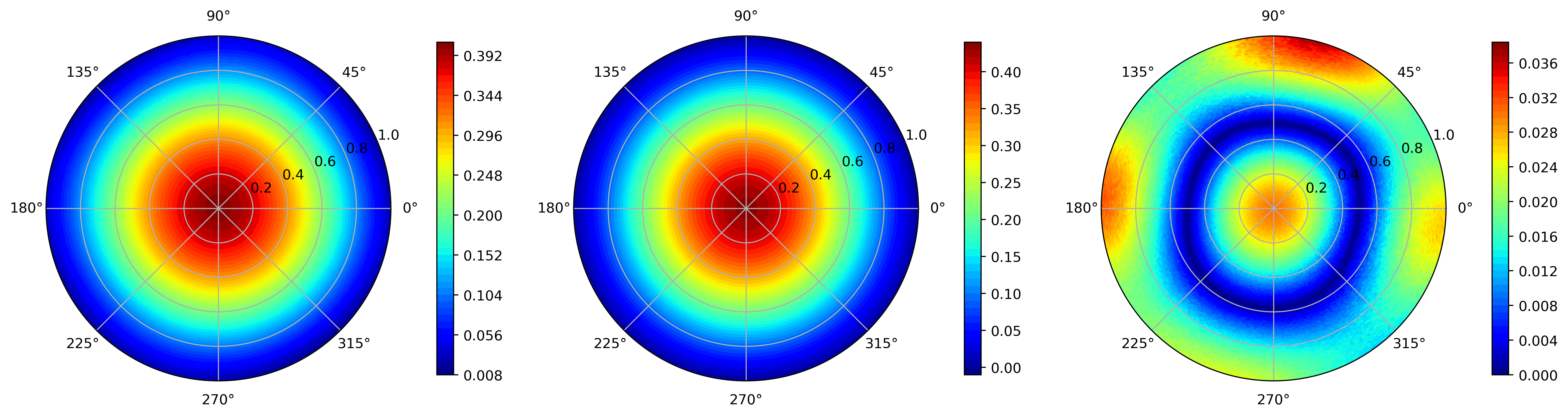}
    \end{minipage}

    \begin{minipage}{0.01\linewidth}
        \centering
        \rotatebox{90}{\scriptsize $x=0$}
    \end{minipage}%
    \begin{minipage}{0.93\linewidth}
        \centering
        \includegraphics[width=0.9\linewidth]{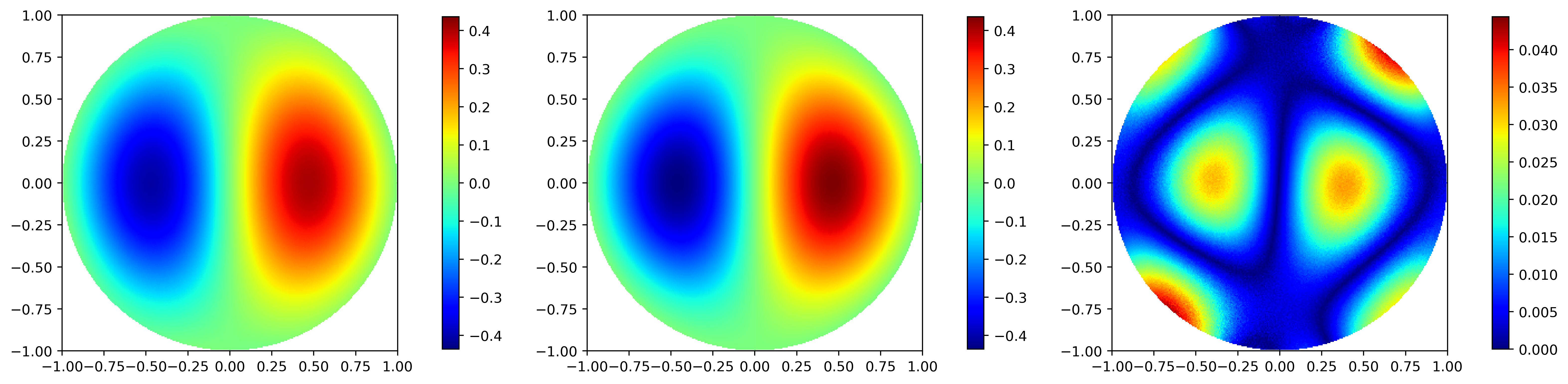}
    \end{minipage}

    \caption{Combined results for the 3D Helmholtz experiment.}
    \label{fig:3d_results_stacked}
\end{figure}

\begin{figure}[ht]
    \centering
    \subfloat[3D Axis Plot]{
        \includegraphics[width=0.45\linewidth]{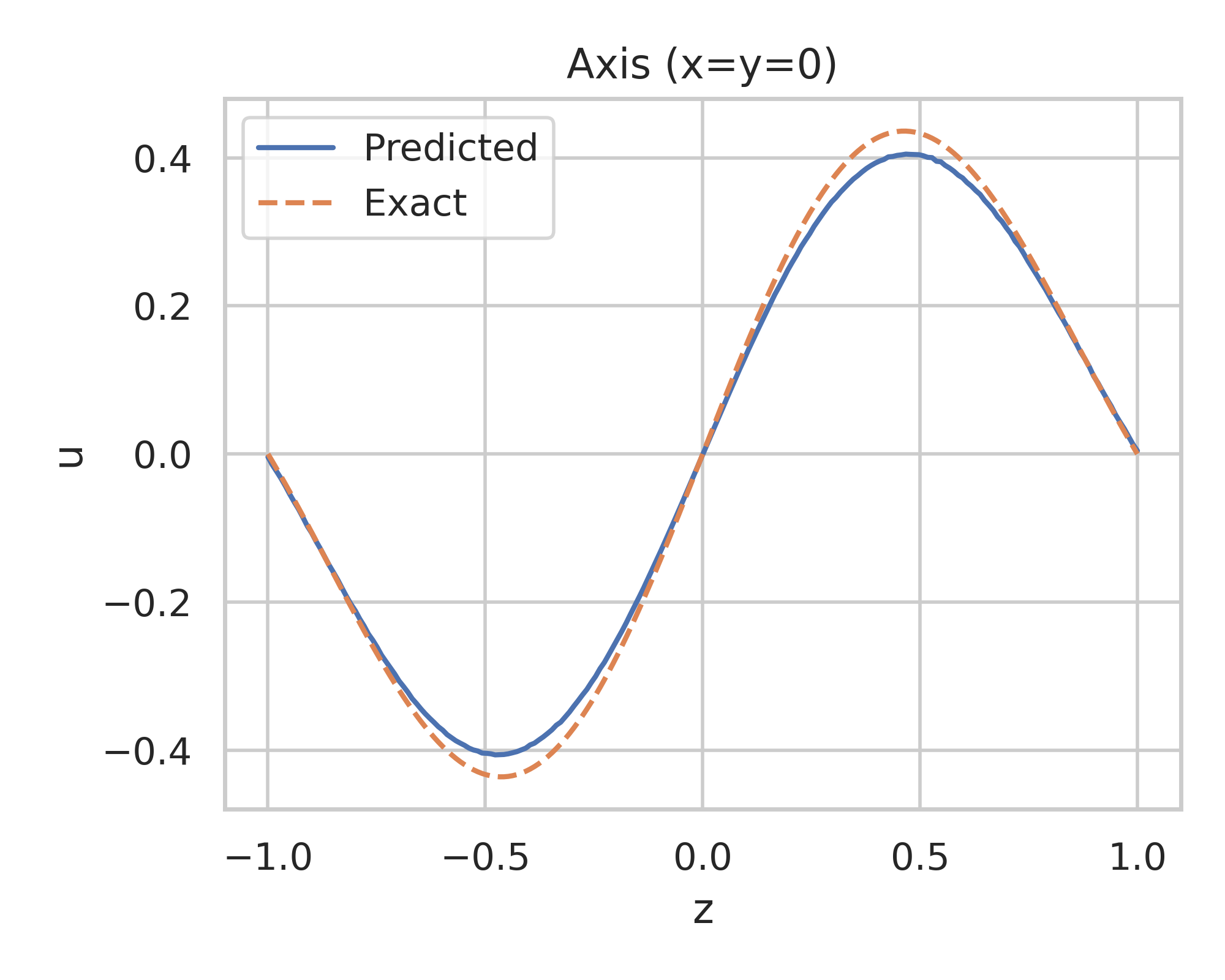}
        \label{fig:3D_Axis}
    }\hfill
    \subfloat[3D Histogram]{
        \includegraphics[width=0.45\linewidth]{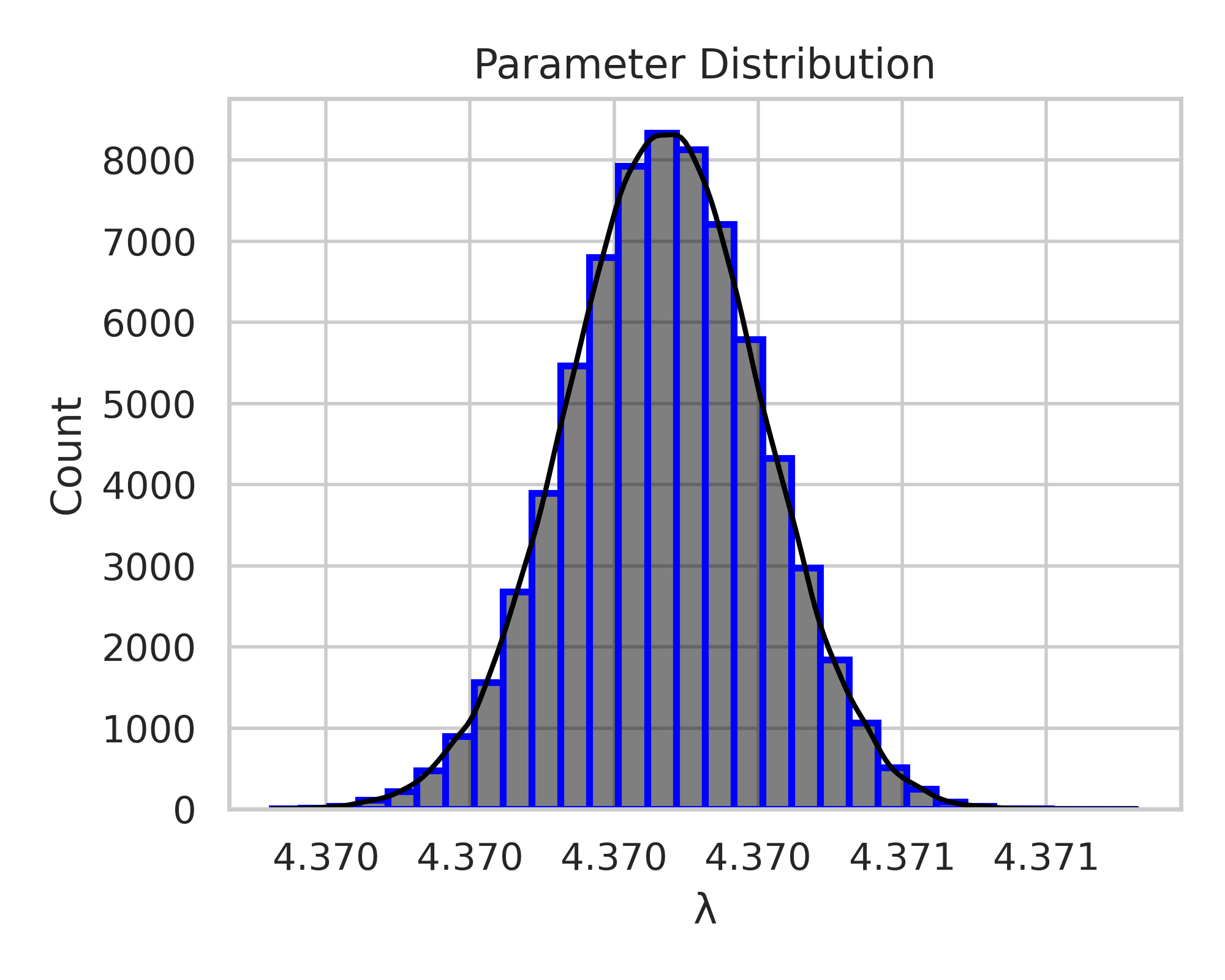}
        \label{fig:3D_Histogram}
    }
    \caption{Axis and Histogram plot for the 3D Helmholtz equation in \cref{eq:3d-eig-problem}}
    \label{fig:3d_axis_histogram}
\end{figure}

\Cref{fig:3d_results_stacked} presents a comprehensive visualization of the model's performance on the 3D Helmholtz eigenvalue problem using both equatorial slices at fixed values and a meridional slice at $x=0$ In each slice, the model's predicted field (left), the exact solution (center), and the point-wise absolute error (right) are shown. The polar plots for $z= 0.0,0.25$ and $z=0.50$ illustrate that the model learns the correct dipole structure expected. The meridional slice, shown in the bottom row, further verifies the agreement between predicted and exact solutions in the full 3D domain. The errors are highly localized and remain within a small magnitude across the domain. To further assess the model's accuracy along a 1D cut, \Cref{fig:3D_Histogram} shows the predicted and exact values of the solution along the central vertical axis $(x=y=0).$

Complementing this, the \Cref{fig:3D_Axis} histogram of the learned eigenvalue parameter $\lambda$  confirms sharp convergence to the ground truth value $\lambda\approx 4.493$, validating that the network accurately identifies the underlying spectral structure.

\section{Conclusions}
\label{sec:conclusions}

This study introduces Bayesian Deep Operator Networks (DeepBayONet) integrated with Physics-Informed Neural Networks (PINNs) to address parameter estimation and uncertainty quantification challenges in solving partial differential equations (PDEs). By employing a Bayesian framework with variational inference, the approach effectively captures both data and model uncertainties, offering robust predictions in noisy and data-scarce environments. The method demonstrates versatility across tasks, including regression and PDE problems in one, two, and three dimensions. 

The framework provides several advantages. It achieves comprehensive uncertainty quantification, enabling reliable predictions under complex scenarios, and maintains competitive accuracy. Additionally, the architecture integrates seamlessly with modern deep learning frameworks, allowing GPU-accelerated applications.

\begin{highlightblue}
The method has limitations that may be addressed in future directions: Careful tuning of hyperparameter can be achieved algorithmically by adaptive strategies for loss weighting. Further refinements that can use non-Gaussian priors to incorporate domain-specific knowledge could enhance parameter estimation and uncertainty quantification. Finally, although \ourmethod{} discovers the parameters of the PDE from the available data and physics, it does not learn a parametrized solution operator of the PDE. We leave that extension to future endeavors.  
\end{highlightblue}

 \section*{Acknowledgments}
 We thank the reviewers for their careful reading and thoughtful feedback, which helped us improve this work.
This research was supported by the CSML lab and the College of Engineering at California State University Long Beach. The work used Jetstream2 at Indiana University through allocation CIS230277 from the Advanced Cyberinfrastructure Coordination Ecosystem: Services \& Support (ACCESS) program, which is supported by National Science Foundation grants \#2138259, \#2138286, \#2138307, \#2137603, and \#2138296.

\bibliographystyle{elsarticle-num} 
\bibliography{references,cpinn}
\end{document}